\documentclass{article}
\usepackage[a4paper]{geometry}
\usepackage[english]{babel}
\usepackage[T1]{fontenc}
\usepackage[utf8]{inputenc}
\usepackage{amsmath}
\usepackage{authblk}
\usepackage{caption}
\usepackage{subcaption}
\usepackage{csvsimple}
\usepackage{booktabs}
\usepackage{adjustbox}
\usepackage{tabularray}
\usepackage{multirow}
\usepackage{colortbl}
\usepackage{xltabular}
\PassOptionsToPackage{hyphens}{url}\usepackage[colorlinks=true,linkcolor=black,anchorcolor=black,citecolor=black,filecolor=black,menucolor=black,runcolor=black,urlcolor=black]{hyperref}
\usepackage{abstract}
\usepackage{cleveref}
\usepackage{natbib}
\usepackage{ulem}

\RequirePackage{etoolbox,xpatch}
\RequirePackage{ifxetex,ifluatex}
\RequirePackage{setspace}
\RequirePackage[table]{xcolor}
\RequirePackage{graphicx}
\RequirePackage{geometry}
\RequirePackage{rotating}
\RequirePackage{titlesec}
\RequirePackage{fancyhdr}

\RequirePackage[marginal]{footmisc}
\RequirePackage{enumitem}
\RequirePackage{authblk}
\RequirePackage{natbib}
\RequirePackage{textcomp}
\RequirePackage[absolute]{textpos}
\RequirePackage[left, pagewise]{lineno}

\definecolor{Gray}{rgb}{0.501,0.501,0.501}

\newcolumntype{L}{>{$}l<{$}} 

\title{Cheap Learning: Maximising Performance of Language Models for Social Data Science Using Minimal Data}
\date{}

\author[1, 2, $\dagger$]{Leonardo Castro-Gonz\'alez}
\author[1]{Yi-Ling Chung}
\author[1, 3]{Hannah R. Kirk}
\author[1]{John Francis}
\author[1]{Angus R. Williams}
\author[1]{Pica Johansson}
\author[1, 4]{Jonathan Bright}

\affil[1]{Public Policy Programme, The Alan Turing Institute, British Library, 96 Euston Rd, London, NW1 2DB UK.}
\affil[2]{School of Geographical Sciences, University of Bristol, University Road, Bristol, BS8 1SS, UK}
\affil[3]{Oxford Internet Institute, University of Oxford, 1 St Giles, Oxford, OX1 3JS, UK}
\affil[4]{Zug Institute for Blockchain Research, University of Lucerne, Luzern, Switzerland}
\affil[$\dagger$]{corresponding author: \url{leonardo.castrogonzalez@bristol.ac.uk}}

\begin{document}
\maketitle 

\begin{abstract}

The field of machine learning has recently made significant progress in reducing the requirements for labelled training data when building new models. These `cheaper' learning techniques hold significant potential for the social sciences, where development of large labelled training datasets is often a significant practical impediment. In this article we review three `cheap' techniques that have developed in recent years: weak supervision, transfer learning and prompt engineering. For the latter, we also review the particular case of zero-shot prompting of large language models. For each technique, we provide a guide of how it works and demonstrate its application and the presence of systematic biases across two different and realistic social science tasks paired with three different dataset makeups. We show good performance for all techniques and we demonstrate how prompting of large language models can achieve high accuracy at very low cost, but biases must be considered.
\\

\end{abstract}

\section{Introduction}
\label{sec:intro}

The systematic analysis of large-scale text data is a growing part of sociology and social science research. The rapid digitisation of social life has meant that daily activity has increasingly left digital `traces' \citep{boyd_six_2011}. Across the social sciences, many of the core objects of disciplinary study can be interrogated through analysis of large datasets created by these traces. Examples abound: large-scale corpora of news articles; datasets of political manifestoes and speeches; textual interactions between individuals around romantic relationships, job searches, discussions around health and climate change, to name but a few \citep{scharkow_measuring_2011, burscher_using_2015, beigman_klebanov_automated_2020, gangula_detecting_2019, souma_enhanced_2019, stede_climate_2021, jensen_language_2021}.

In the past, large-scale text corpora were addressed through equally large-scale manual coding efforts (see, for example, \citet{lehmann_manifesto_2023}), whereby texts are reviewed by teams of trained coders. However, around a decade ago, work emerged pointing to the potential for automatic content analysis to shed light on the social dynamics contained within these datasets, making use of insights from the field of machine learning \citep{wiedemann_opening_2013, grimmer_text_2013, grimmer_machine_2021}. By now a considerable body of work has developed methods for automating content analysis in the social sciences \citep{nelson_small_2018, nemeth_machine_2020, lichtenstein_contextual_2021}, which offers obvious potential benefits in terms of saving time and effort, and thus allowing for analyses at a scale that would be beyond most teams of manual coders \citep{burscher_teaching_2014}. 

However, while the potential opened up by this automatic work is important, there remains a weakness in the automatic content analysis based on the `conventional' paradigm of machine learning (as we will refer to it here): it still in reality contains a considerable manual component. As we will describe more fully below, in order for a classifier to be trained to identify a given concept, labelled training data is required, often in large quantities. Hence a significant time investment is required before such a classifier can be developed. Whilst such upfront work can be justified in industry contexts that may require millions of data points to be labelled automatically on an ongoing basis, many analytical research tasks in the social sciences could actually be achieved with this initially labelled dataset, rendering the training of the model somewhat irrelevant. Hence, the practical advantages of this type of machine learning approach in the social sciences are not always clear cut, with some saying that the time and financial savings from automated methods are exaggerated \citep{de_grove_what_2020, nemeth_machine_2020} and others preferring methods granting researchers greater measurement validity \citep{baden_three_2022}. 

The heavy data requirements of conventional machine learning approaches have led to considerable research interest in techniques that might reduce data requirements without sacrificing performance or analytical rigour \citep{al-jarrah_efficient_2015, goldsteen_anonymizing_2022, galal_federated_2024}. In computer science, such techniques are now developing very actively, with considerable advances and a wide array of new techniques developing. However, only a few papers have emerged trying to make these techniques accessible to social science researchers \citep{molina_machine_2019, macanovic_text_2022, wankmuller_drawing_2022, costantini_high-dimensional_2023}, and the uptake remains patchy at best. \\

The aim of this article is to remedy this deficit and offer an introduction to these `cheap' learning techniques, as compared to their more expensive conventional counterparts. We discuss three main approaches. First, we look at weak supervision, a method for formalising background knowledge of a subject into a systematic approach to data labelling, making use largely of keyword-based approaches. While it might not be the technique that offers the highest performance, it is transparent and easy to implement, retaining the simplicity of the dictionary technique but assuaging some of the criticisms of how dictionaries are both arbitrary and difficult to validate \citep{grimmer_text_2013}. Second, we discuss transfer learning, a technique which involves taking a model trained in one context and applying it to another, making use of a minimal amount of retraining data. We show how good results can be achieved with small amounts of novel data by building on pre-existing models. Finally, we discuss `prompt engineering': prompting large language models (LLMs), such as the GPT family created by OpenAI, as a means of content classification. We look at both direct use of generative models for such classification (sometimes called `zero-shot' learning), and also how these techniques can be combined with some elements of transfer learning. LLMs in particular have generated much excitement as a very cheap method of content classification \citep{gilardi_chatgpt_2023}; however, we show that at the moment they do not outperform more traditional approaches for all tasks. 

Each one of these techniques offers the potential for realising the benefits of supervised machine learning, but at less cost in terms of human labelling, hence offering the tantalising potential of finally realising the potential of automatic content analysis in sociology and the social sciences in general. For each one of these techniques, we describe how it functions, discuss approaches to validation, and discuss common potential use cases. In an empirical section, we then illustrate using existing annotated datasets how good performance can be achieved with a minimal amount of labelled data (up to 1024 data points), as well as showing how performance improvements relate to increases in labelled data. We also compare all of these techniques to the Na\"ive Bayes approach and to a logistic regression classifier, two classic methods from conventional machine learning \cite{10.1007/978-3-540-30549-1_43, aborisade_classification_2018}. Although these methods were developed more than two decades ago, their simplicity of training and deployment makes them a good baseline to compare with newer and more sophisticated methods in a low-resource context.

We complement our performance analysis with an additional experiment to study the biases each technique can have in classifying text for each task. While the most recent OpenAI's GPT models have been used to annotate text \citep{gilardi_chatgpt_2023}, the potential systemic biases they have at an annotation task can lead to incorrect results \citep{ashwin_using_2023, spirlingWhy2023}. With this bias analysis, closed models like OpenAI's GPTs can be compared to more classical machine learning techniques in another useful dimension for social scientists besides their performances. 

We should note the existence of a handful of other papers that have made similar contributions to our own \citep{terechshenko_comparison_2020, treviso_efficient_2023}. Whilst these are valuable works, our paper is distinctive in a number of ways. First, these papers address one technique exclusively, whereas ours compares three at the same time. This comparison allows us to explore the `labelling budget' (the amount of manually labelled data required for each technique), as well as discussing how different methods perform well in different circumstances. Second, our analysis is not limited to the performance of labelling data but also studies the systematic bias of each technique. Finally, alongside our paper we release code that will allow other researchers with a practical working knowledge of the Python language to implement these techniques in their own research\footnote{\url{https://github.com/Turing-Online-Safety-Codebase/cheap_learning}}. 

With this paper, we contribute by giving sociologists and social scientists tools and strategies to address challenges where data scarcity is found. By offering available online resources and a performance and bias analysis done on two well-known examples in the literature \citep{maas_learning_2011, wulczyn_ex_2017}, we hope for social scientists to apply these strategies in key challenges where data scarcity and bias might be embedded into the problem, like interview analysis of under-represented groups \citep{deterding_flexible_2021,  Small_qualitative_2022, bonikowski_ends_2022, ashwin_using_2023, atari_which_2023} or text analysis and labelling using historical text archives \citep{jaillant_archives_2022, marciano_archival_2018, rahal_rise_2024}. \\


While acknowledging the importance of unsupervised machine learning approaches \citep{hutto_vader_2014, shelke_efficient_2022}, we chose not to work with these techniques. Given that unsupervised learning discovers structure in data rather than labelling according to prior theoretical concepts of interest, we consider unsupervised learning techniques working fundamentally different to the features of supervised learning we are interested in testing, thus falling out of the scope of this study.
In the same way, we are also not including open-source large language models (LLMs) for this study. On the one hand, we are interested in showing the monetary costs, performance and potential biases of using the API of closed models like the OpenAI's GPT-3.5 (ChatGPT) and GPT-4.0, which have grabbed much attention \citep{kalla_study_2023} in the academic and non-academic world. On the other hand, we decided to centre our attention on the mentioned closed LLMs to avoid this study becoming another comparison between the different LLMs now available for researchers \citep{yu_open_2023}. \\

Our paper is structured in the following way. In section \ref{sec:lit_review} we give a general overview of how we define cheap learning techniques, and also justify our focus on the techniques we are looking at. Then, we provide a guide to each technique, describing how they function, discussing approaches to validation, and discussing common potential use cases. In section \ref{sec:methods} we describe our approach to the empirical illustration of the models and describe the data we make use of. Finally, in section \ref{sec:results}, we illustrate the use of these techniques using existing annotated datasets to showcase how good performance can be achieved with a minimal amount of labelled data (as well as showing how improvements in performance relate to increases in labelled data). We also discuss concerns around balanced and unbalanced data, showing how these might affect preferences for models. We conclude by providing perspective on directions in the field, especially in the light of increasingly powerful and available large language models. 
\section{Cheap Learning Techniques}\label{sec:lit_review}

In this paper, we address methods for automating content analysis in the social sciences using techniques from the field of machine learning. Our focus is on algorithms that can be implemented `cheaply' when compared to what we will define as the `conventional' approach to automatic content analysis, which involves training a model from scratch with a large labelled training dataset (though we refer to it as `conventional', we should also say that it is an approach that is less and less the norm within computer science). As highlighted in the introduction, we address three main approaches: weak supervision, transfer learning and prompt engineering. We regard these algorithms as `cheap' to implement because they typically require much less labelled data to achieve good performance than the conventional approach (though all of them still require some such data, particularly for validation). In this section, we will provide an overview of the three techniques we are introducing in the paper. For each technique, we provide a general overview of the history and motivation, and provide a guide on how to implement it, including notes on how to approach evaluation of the technique. We also reflect on the cost of each algorithm, and the amount of computing power required to implement it (with the proviso that, of course, computing power increases at a regular pace). Finally, we provide our thoughts on how each technique might adapt to different problem sets in the social sciences.

We should note that we do not address `unsupervised' machine learning, a technique which involves no manual labelling (at least in the input stage). We omit this technique here partly because it has already been covered by a relatively extensive set of social science applications (e.g. \citet{berry2019supervised, molina_machine_2019, Waggoner_2021}), and also because we do not consider it a direct competitor to supervised machine learning. While unsupervised learning offers the potential to group text documents into lexically similar sets (often referred to as `topic modelling') it does not provide the direct link to theoretical categories of interest which is offered by supervised machine learning: that is to say there is no guarantee that identified topics will map neatly to theoretical concepts and hence be directly applicable to an analytical social science question. Therefore, while the technique is an excellent approach for exploratory analysis of a large dataset, the lack of ability to define concepts at the outset can make it challenging to use topic models for explanatory hypothesis testing (though we do note the emergence of seeded topic models \citep{doi:10.1177/08944393231178605} as a potential partial remedy to this problem), and of course post-hoc labelling of topics can allow them to be built into an explanatory framework. 

We contrast all of these approaches to the conventional approach to text classification through machine learning, which is worth briefly reviewing here (the subject has already been treated extensively in, for example, \citet{kotsiantis2006machine}, \citet{jordan2015machine}, and \citet{mahesh2020machine}). This approach seeks to produce a model that, given an input text, estimates the probability that this text belongs to a given analytical category. Such models are based on input data: a corpus of text that has been labelled by hand such that each piece of text is assigned to a category. One classic example, that we make further use of below, would be to label a given piece of text as containing `hate speech' or not, to produce a text-based classifier that can automatically detect hateful text. 

The corpus of text data is typically divided into three groups: training data, development data, and test data. Training data are the data used to train the model to begin with: the model's decision making will be based on patterns learnt from these data. There are a wide variety of approaches to learning a model from training data that we will not seek to review here (good overviews are provided by \citet{kotsiantis2006machine, cherkassky2007learning, jordan2015machine, mahesh2020machine, gupta2020survey}). Broadly speaking, they will rely on identifying differences in linguistic patterns between the different classes of data for which a classification is sought (for example, hateful text is likely to contain different words to non-hateful text). Development data are data used during the process of model training to provide an estimate of the model's performance (with a variety of metrics typically calculated for this evaluation - see \citet{fatourechi2008comparison, hossin2015review}). These development data allow for decisions to be made during the process of model training, for example selecting appropriate hyperparameters \citep{luo2016review, 50c707ab9d8b428d85ee3689b84697a8, KANDEL2020312}. Test data, finally, are used to make an assessment of model performance once the process of training is complete. Test data are typically kept separate until the final model evaluation to guard against `overfitting', a phenomenon whereby (through repeated trial and error with different models and different hyperparameters) a model is developed that performs well on the development dataset but generalises poorly to unseen data. 

When implementing such a process within the social sciences, the amount of data required (the `labelling budget', \citet{lindstrom2013drift, 10.1145/3514221.3517904}) is the core practical concern. High-quality labelled data is expensive and time-consuming to generate. Firstly, coders need to be recruited and trained on the concepts of interest, and need to establish a common understanding of the concepts they are coding (often measured through high inter-coder reliability scores). Once this process is complete, labelling needs to take place, which is a time-consuming and repetitive manual task (and gold standard scientific journals often require data to have been coded by multiple coders, \citet{10.1145/3290605.3300637, geva-etal-2019-modeling}). The exact amount of time will of course be task-dependent (varying by, for example, the amount of text that needs to be reviewed, the amount of potential classes, and the complexity of the coding task). However, there is a general expectation that the process will be time-consuming, driven by the fact that the data requirements of high-performance classification models are typically large \citep{sun2019fine, zheng2020effects, anaby2020not}, and that increasing the size of the training dataset is the standard recommendation for improving performance \citep{shams2014semi, 10.5555/3455716.3455856, hsieh-etal-2023-distilling}.

The approaches we review below are all motivated by a desire to reduce the cost of the labelling budget to produce a high-performing model, particularly the training and development datasets (a high-quality test dataset for evaluation remains essential to the process however). Each of the techniques approach this task in a different way: weak supervision does not necessarily reduce the size of the labelling budget, but makes large training datasets cheaper to put together; transfer learning seeks to leverage other pre-existing models to reduce the requirements for novel labelled data; while prompt engineering seeks to exploit `emergent' properties of large language models that have been trained for generic language production tasks but can also be leveraged for classification. 

As a core element of our paper is cost, we should also mention some other costs that are relevant here beyond those needed for the creation of labelled data. There are hardware costs implied in some of our work below: though much of what we describe can be implemented on a generic personal laptop, some may require the use of an institutional or cloud server. For commercial language models, there are also costs in terms of API access that could mount up if there are requirements to label large volumes of data. Finally, there are some time costs required for training models, that can mount up if a long cycle of experimentation is required to find the ideal configuration of hyperparameters. We will also comment on these costs for all of the techniques we discuss.   

\subsection{Weak Supervision}

Proposed in \citet{ratner_data_2016} and extended in \citet{ratner_training_2019}, weak supervision is an approach to text classification that seeks to cheaply build labelled data using heuristic rules based on pre-existing background knowledge of the concepts being labelled. Data labelled in this fashion can either be applied directly to an analytical task, or leveraged to build a further classification model. 

As a simple example of weak supervision, we could imagine a task based on classifying international news articles based on which country they relate to. A first step to build a model through the conventional machine learning approach would be to take a dataset of news articles and label by hand the country to which they relate (conscious that an article might relate to more than one country). However, this approach essentially ignores background knowledge that will be largely obvious to people with domain expertise: we know that (for example) an article containing the keyword `Canberra' is most likely about Australia, whilst one containing the keyword `Boston' likely refers to the US. Integrating this knowledge would seem like a valuable efficiency. 

Using keywords to classify content in this fashion, sometimes called the `dictionary' approach, does of course have a long history in the social sciences and indeed many early approaches to content classification were based on keywords \citep{hulth-megyesi-2006-study, onan2016ensemble}. However, such approaches received strong criticism as being hard to evaluate properly \citep{grimmer_text_2013}. In the absence of manually labelled data, it was hard to measure common evaluation metrics such as the precision and recall\footnote{Precision refers to the number of true positives divided by the total amount of the positive elements retrieved. In other words, precision measures the reliability of the results. On the other hand, recall refers to the ratio of the true positives divided by the total amount of relevant items (true positives + false negatives).}, or understand how many false positives and negatives might be generated (for example, an article on Boston, Lincolnshire in the UK would be obviously misclassified by the above approach).

Weak supervision can be viewed as a method that seeks to circumvent some of these difficulties: benefiting from the simple but powerful domain knowledge that experts can bring to bear on classification, without sacrificing the rigor or evaluation metrics used in more conventional machine-learning approaches. It has already been applied in a wide variety of different contexts like text data in the fields of medicine and health sciences \citep{Fries2021, Silva_2022}, chemistry \citep{Mallory_2020}, detection of fake news \citep{Shu_2021}, and more generally to study sentiment analysis \citep{Jain_2021}, and ontology and computational linguistics problems \citep{Maresca_2021, Berger_2021}. 

Weak supervision is based on two main components: labelling functions (LFs) that encode background knowledge into formalised rules; and a probabilistic model that is trained with the noisy, overlapping labels obtained from applying the functions to the labelled data set, and its respective prior labels. 

Labelling functions are abstract rules that take an unlabelled data point and either assign a label if a criterion is met or abstain from assigning it otherwise. The presence or absence of keywords contained in text data is one of the most obvious types of such rules, however, there is no need to only make use of keywords. Regular expressions, Named-Entity Recognition (NER) systems, results from other models such as sentiment classifiers, the length of the document to be classified, or its source; any of these document features can be incorporated into a labelling function. This diversity of potential criteria makes it possible to have topic-specific sets of labelling functions obtained from the particularities of the topic to be classified, going a step beyond the dictionary approach. Some examples of simple labelling functions that we make use of in the experimental section of the paper are presented below. For further examples, the list of all the labelling functions used for the classification tasks presented in the paper can be found in the Appendix \ref{ssec:labelling_functions} of the online supplement.

\begin{quote}
\begin{itemize}
    \item LF1: \texttt{return NEGATIVE if polarity(text) < -0.25 else ABSTAIN}
    \item LF2: \texttt{return POSITIVE if regex found in text else ABSTAIN} 
    \item LF3: \texttt{return ATTACK if length(text) > LIMIT else ABSTAIN}
\end{itemize}
\end{quote}

In the examples above, variables \texttt{NEGATIVE} and \texttt{POSITIVE} refer to the classes used for the binary movie sentiment classification exercise, while the \texttt{ATTACK} refers to one of the classes used for the Wiki Personal Attacks exercise explained below. In the same way, the variable \texttt{ABSTAIN} refers to the value that the function returns when it abstains from classifying a text. The polarity function refers to the sentiment analysis function to measure the polarity of a given text between -1 and 1. \texttt{regex} is used in this example as any regular expression that would refer to a non-abusive text, e.g. ``This is such a good idea!''.
LF3 is a good example of how domain knowledge can be used in weak labelling. As seen in the Wiki Personal Attacks exercise \citep{wulczyn_ex_2017}, cyberbullying can take many forms \citep{sathyanarayana_rao_cyberbullying_2018}. One of them is to repeat a sentence or a set of words in a comment until the maximum number of characters allowed is hit. LF3 accounts for that, labelling unusually long texts as personal attacks by measuring the length of the text and checking if it is greater than the maximum length allowed (\texttt{LIMIT}). 

The exploration subset of the dataset serves to fine-tune or debug the set of labelling functions. Weak supervision enables measuring \textit{coverage} and \textit{overlap} of these functions in the different sets (details of the LFs used in this work are presented in Appendix \ref{sec:appendix} of the online supplement). Coverage for a specific labelling function indicates the percentage of data points it labels without abstaining. On the other hand, overlap represents the percentage of data points where two or more labelling functions do not abstain. Ideally, for a given class, we aim for a set of labelling functions that collectively cover the entire exploration set without significant redundancy in their actions, measured by a high overlap.

Once the labelling functions are created, they are then applied to the training set, generating a label matrix. Each row in the matrix represents a data point, and each column represents a labelling function. Since the training data is already classified, this label matrix is used to compute weights for each labelling function using a Bayesian probabilistic model, called a \texttt{LabelModel} by the authors \citep{ratner_training_2019}. 
The overall model, incorporating the labelling functions and their respective weights, is then used on the test set for classification. 

The outcome of a weak supervision process will be a partially labelled dataset (as we expect the labelling functions to abstain in some cases rather than suggesting a label). Depending on the coverage achieved, the partially labelled set may already be enough to answer the question at hand. If it is not, a common next step would be to train a conventional machine-learning model to label the remaining data points. The success of the approach will depend on whether the partially labelled data is good enough to produce a strong classification model, though as we will present below the approach shows good potential for a number of tasks. 

In addition to its very transparent and simple approach to producing labelled data, a further advantage of weak supervision lies in its lack of significant computing costs. Developing and applying labelling functions are computationally cheap tasks that can be tackled on any standard consumer laptop and labelling even a very large dataset in this fashion would take little time or processing power. 

\subsection{Transfer Learning}

The approach to conventional machine learning described above starts from a blank page. Given a classification problem, and a large enough pre-existing labelled dataset, a model is trained from scratch, experimenting with different hyperparameters, trying to achieve the best possible accuracy on a development dataset. The final evaluation score is then calculated on previously unseen test data. 

Transfer learning starts from a different premise \citep{5288526, weiss2016survey, zhuang2020comprehensive}. Rather than starting from scratch, it looks for pre-existing models (potentially from closely related source domains), and seeks to leverage the information already contained within them \citep{weiss2016survey, neyshabur2020being, bashath2022data}. A model built on this approach will still require some novel training and development data; but far less than in the conventional approach. Furthermore, given the variety of input data, transfer learning also offers a method for obtaining models with better generalisation properties. These advantages mean that transfer learning is a rapidly growing field of work \citep{NEURIPS2019_c04c19c2, chung2020italian}. Transfer learning has shown great success in several real-world applications, such as text classification \citep{10.5555/3455716.3455856, ali2022hate}, image recognition \citep{Guo_2019_CVPR, kolesnikov2020big}, and natural language generation \citep{10.5555/3455716.3455856, golovanov-etal-2019-large}.  

There are a variety of ways in which a pre-trained model might be selected. Domain specificity is obviously a concern: if seeking to build a classification model for hate speech, selecting a pre-existing hate speech classifier may be a sensible place to start. Beyond this, language ability (e.g., monolingual or multilingual), model architecture (e.g., model size, number of parameters and speed), computation power (e.g., hardware and number of GPUs required) and model performance (e.g., accuracy) all might be a consideration. With recent advancements in natural language processing, transformer-based large language models that have been developed and pre-trained on massive corpora (e.g., Common Crawl \citep{10.5555/3455716.3455856}) using various pre-training objectives \citep{devlin-etal-2019-bert, 10.5555/3454287.3455457} are becoming more and more popular as the starting point of transfer models focused on text classification. These general-purpose models lack any domain specificity, but they have nevertheless demonstrated strong performances and can leverage knowledge from source data to downstream tasks through fine-tuning methods \citep{pmlr-v97-houlsby19a, peters-etal-2019-tune}. As a general rule larger models mean better performance whilst requiring high computation power and memory consumption. A key question for many in the social sciences will be whether the model being selected can be run on a researcher's laptop or whether a cloud computing platform that can offer greater performance (such as AWS, Azure or GCP) will be required \citep{pierleoni2019amazon, kamal2020highlight}. However the technology here is evolving constantly, and the possibility of running ever larger models on contemporary consumer hardware is always increasing \citep{pierleoni2019amazon}. 

Once a base model has been selected, it is then fine-tuned on the training set. The ideal size for a training set is not easy to know in advance, but in the experiments section below we will present a few rule of thumb examples for the ideal size. This fine-tuning has the aim of improving model performance for the specific task at hand. There are a variety of different ways to approach this task, tuned through hyperparameters, and many of these have conventional values that have been found to work reasonably well in a variety of settings. Training epochs indicate the number of times a model sees or passes over the whole training dataset, often in the range between 1 and 5. Learning rate refers to how much to adjust the model (i.e. model weights) with respect to the knowledge learned each time the model weights are updated \citep{bengio2012practical}, with 0.0001 and 0.0003 sometimes suggested as working well for many problems.\footnote{See for instance: \url{https://huggingface.co/docs/transformers/model_doc/t5}} A learning rate that is very large could make rapid divergence of model weights, likely resulting in unstable training and a sub-optimal solution, whereas a learning rate that is too small may require longer training time for models to converge or arrive only at a local minimum. Batch size decides the number of instances processed before the model weights are updated, which is generally a power of 2 in the range between 8 and 128. While increasing batch size generally requires less training time, it consumes more memory and may not lead to high accuracy \citep{50c707ab9d8b428d85ee3689b84697a8, KANDEL2020312}. A random seed will also indicate the initial state of the model parameters.

Finally, the finished model can be evaluated to assess if a model is learning as expected (e.g., whether the model is overfitting or not). This includes applying the trained model to the test set using various metrics, including precision, recall, F1, and accuracy. Graphical representation is another helpful way of diagnosing model performance, for instance, through learning curves of model performance over the training process, confusion matrix and receiver operating characteristic (ROC) curve. These methods are the same as those used in the conventional approach described above, hence we will not describe them in detail here. 

\subsection{Prompt Engineering}

The final technique we will introduce is prompt engineering, also known as in-context prompting or prompt learning. This is a technique for guiding large language models to perform a task (e.g., classification) via textual prompts \citep{10.1145/3560815}. Training a prompt model is similar to transfer learning in a way, in that we are exploiting pre-existing knowledge within a model. However, unlike the above approach, we no longer seek to train a model in a classic sense, but rather exploit prompts as guidance to elicit the generative potential already held within large language models to generate answers to classification questions. 

In theory, a nuanced prompt tailored to a task can produce high performance in a classification task even more cheaply than transfer learning. With handcrafted or automatically generated prompts, prompt engineering is particularly beneficial for scenarios/domains where (1) language models show strong performance and (2) labelled examples are scarce or not available \citep{schick-schutze-2021-exploiting, 2021arXiv210310385L, hu-etal-2022-knowledgeable}. Such prompting has been explored and has shown promising success in several domains, such as text classification \citep{10.5555/3455716.3455856, ali2022hate}, image recognition \citep{Guo_2019_CVPR, kolesnikov2020big}, and natural language generation \citep{10.5555/3455716.3455856, golovanov-etal-2019-large}. 

Unlike transfer learning, in prompt engineering a language model is instructed to perform with the help of task description as an auxiliary training signal using `pattern-verbalizer pairs' \citep{schick-schutze-2021-exploiting, schick-schutze-2021-just}. A pattern is a natural language prompt or an expression that transforms inputs into cloze-style tasks (i.e. a portion of a passage is removed or closed, and the task is to predict the removed chunk that would best fit the surrounding context). A verbalizer refers to mapping a label/class (e.g., movie) to a list of task-representative words, also called label words, (e.g., action, drama and adventure). For example, given a binary movie sentiment classification task (as we explore below), an \textit{example input in italics}, taken from IMDb Movie Review Sentiment dataset \citep{maas_learning_2011}, with a \textbf{\textit{pattern in bold}} is shown as follows:

\begin{quote}
    \textit{Two Hands restored my faith in Aussie films. It took an old premise and made it fresh. I enjoyed this movie to no end. I recommend it to those people who like Guy Ritchie films. Bryan Brown was fantastic and just about perfect in a role tailor made for him. \textbf{It was? Negative or Not Negative} [MASKED]} 
\end{quote}

The answer is masked and converted into a class based on a verbalizer, for instance, `Positive': `good', `great'; `Negative': `bad'. 

Writing a good prompt is crucial for effectively giving instructions and obtaining accurate responses from language models. While interest in prompt design has grown considerably in various domains, such as text classification \citep{10.5555/3454287.3454581, hu-etal-2022-knowledgeable}, and knowledge probing \citep{petroni-etal-2019-language, davison-etal-2019-commonsense}, studies on systematic evaluation of the effects of prompts are limited. We provide several heuristics that can help write a good prompt. First, a prompt is generally clear and specific with relevant background information provided. A prompt can be written in the form of a statement or question. Second, when dealing with complicated tasks or multiple sub-questions, breaking down the tasks into sub-tasks can help models tackle each task one by one \citep{wei2022chain}. In \citet{press-etal-2023-measuring}'s study, they find that asking language models directly how to write a good prompt for a given task of interest and then writing a prompt following what the language model responded can yield better answers. Third, if the response is required to be in a specific format, delineate the format with examples (e.g., make step-by-step reasoning, and provide a pros and cons comparison). Lastly, a good prompt generally benefits from several iterations and experiments by rephrasing the prompt and providing additional information. Much like tuning the hyperparameters in a conventional machine learning model, there is no one formal way of creating the ideal prompt: a degree of experimentation is always required. 
 
Verbalizer design is also worth considering. A verbalizer maps a label to a list of label words which are recommended to have diverse coverage and little subjective bias for eliciting comprehensive semantic information from language models \citep{hu-etal-2022-knowledgeable}. Previous work experimenting with multiple label words for each label shows that the optimal size of label words is generally around three, while varying the number of label words does not lead to significant improvements \citep{schick-etal-2020-automatically, shin-etal-2020-autoprompt}.

Just as with the other techniques above, labelled test data is essential for the evaluation of a model based on prompt engineering. However the amount of training data required is open to question. Up until very recently, prompt engineering models also largely made use of some training data to boost performance through fine-tuning, similar to transfer learning above. But the emergence of openly available large language models such as the GPT family from OpenAI has challenged this paradigm, creating the potential to prompt models without any pre-training at all, sometimes called `zero-shot' learning \citep{DBLP:journals/corr/abs-2109-01652,ratner_training_2019}. As we will show below, performance can be high even in these zero-shot cases. 

All large language models have a set of hyperparameters that also need to be considered when setting them up for a classification task \citep{bommarito2022gpt,rehana2023evaluation}. Probably the most crucial to mention here is `temperature' which affects how deterministic the output of the model is. More specifically, given that the input into a large language model is a sequence of text (sometimes called a `context window'), the output of the model is a list of words with associated probabilities of being the next word in the sequence. The model then picks a word from this list based on the temperature, with a higher temperature making it more likely to pick lower probability words. For creative writing tasks, setting a relatively high temperature seems to generate better performance \citep{10.1145/3180308.3180329, caccia2019language}; for classification tasks, lower temperatures seem like a better place to start \citep{rehana2023evaluation}. However, the `best' temperature is an open question and hard to define in abstract. 

Similar to prompt engineering, zero-shot prompt engineering is particularly beneficial for scenarios where (1) language models show strong performance and (2) labelled examples are scarce or not available \citep{schick-schutze-2021-exploiting, 2021arXiv210310385L, hu-etal-2022-knowledgeable}. Furthermore, while zero-shot prompt engineering can benefit from narrow prompting aimed at a specific task \citep{10.1145/3560815}, broader prompting aimed at eliciting wider cognitive abilities in large language models has also seen success \citep{kojima2022large}. 

A few final comments on other resources required for prompt engineering are worth mentioning. While developments in the area of model miniaturization are continuing rapidly \citep{bai_towards_2022}, at the time of writing running the latest generation of large language models on a consumer laptop was not a realistic possibility, hence a server running a GPU was still likely a requirement.\footnote{Statements on the size of models that a typical laptop can run will go out of date quickly, however at the time of writing Llama 2-70b, arguably the best open source LLM, required about 320 GB of RAM to run \citep{noauthor_sizing_nodate}, which is beyond the realistic range of a typical laptop. Running quantized versions of Llama is undoubtedly possible, though quantized versions may offer lower performance as well as still being quite slow to run.} LLMs are also accessible through APIs, however as we will describe below these often have a cost attached to them.
\section{Methods}\label{sec:methods}

Having introduced the three cheap learning techniques under study in this article, in the remaining part of this paper we want to provide an illustration of them in action, by applying them to two example binary classification tasks related to personal abuse and the sentiment of a movie review. We have chosen these two tasks because we believe they are relatively standard examples of the types of classification task that are appropriate for the social sciences.

In addition to providing an illustration of the techniques in action, our experiments will also help to explore the extent to which these techniques are truly `cheap', by exploring the amount of training data required to achieve different levels of performance, which we refer to as the labelling budget (noting again that for all of the methods we discuss there would still be a requirement to develop a test dataset). 

In this methods section, we describe our approach to these experiments: we look first at the datasets we have chosen and how we sample from them, then describe the classification tasks we seek to perform, then discuss the computing infrastructure used for the experiments, and also explore the particular hyperparameters we have made use of in our experiments. 

\subsection{Data}
\label{ssec:data}

In our experimental section we address two classification tasks: binary abuse classification and movie review sentiment classification. In order to conduct experiments on these tasks, we make use of existing publicly available labelled datasets. For the abuse classification task we make use of the `Wikipedia Talk: Personal Attacks' dataset \citep{wulczyn_ex_2017}, containing 115,864 comments from English Wikipedia labelled according to whether they contain a personal attack (no personal attack - 88.3\%, contains personal attack - 11.7\%). This dataset is a popular one among social scientists, already employed in a variety of studies \citep{tay_are_2022, tay_charformer_2022, dinan_anticipating_2021, sarwar_neighborhood_2022, xu_bot-adversarial_2021}. 

For the movie sentiment task, we rely on the `IMDb Movie Review Sentiment' dataset \citep{maas_learning_2011}, containing 50,000 movie reviews labelled according to whether they have a positive sentiment or negative sentiment (negative sentiment - 50\%, positive sentiment - 50\%). The dataset has also been extensively studied in the literature \citep{tay_charformer_2022, priyadarshini_novel_2021, yu_adaptsum_2021, birjali_comprehensive_2021, saunshi_mathematical_2021}.

The two chosen datasets have a number of differences, the most obvious being the context in which text entries were written. The Wikipedia Personal Attacks dataset entries are comments on an article, that could be directed at another user or the article's content, whereas the IMDb Movie Sentiment entries are user reviews, all targeted at some piece of media, usually providing some kind of detailed critique. This difference in context could suggest a greater range of types of responses, and possibly differentiation between classes, in the Wikipedia dataset compared to the IMDb dataset.

This context is reflected in the length of the entries across the two datasets (Table \ref{tab:text_length}), with entries from the IMDb dataset being over three times as long as entries from the Wikipedia dataset on average (1297 characters vs. 402 characters). Furthermore, the balance of positive and negative labels in the datasets represents another difference. While the IMDB Movie Review Sentiment dataset has an even 50/50 split between labels, the Wikipedia Personal Attacks dataset is unbalanced, which provides another dimension to examine performance between techniques over. Evaluating the chosen cheap learning techniques across datasets with these differences provides a more generalisable view into the pros and cons of these techniques across different scenarios a social scientist might encounter.

\begin{table}[h]
    \centering
    \captionsetup{justification=centering}
    \footnotesize
    \begin{tblr}{
      row{2} = {c},
      cell{1}{1} = {r=2}{},
      cell{1}{2} = {r=2}{},
      cell{1}{3} = {c=4}{c},
      cell{3}{1} = {r=3}{},
      cell{3}{3} = {r},
      cell{3}{4} = {r},
      cell{3}{5} = {r},
      cell{3}{6} = {r},
      cell{4}{3} = {r},
      cell{4}{4} = {r},
      cell{4}{5} = {r},
      cell{4}{6} = {r},
      cell{5}{3} = {r},
      cell{5}{4} = {r},
      cell{5}{5} = {r},
      cell{5}{6} = {r},
      cell{6}{1} = {r=3}{},
      cell{6}{3} = {r},
      cell{6}{4} = {r},
      cell{6}{5} = {r},
      cell{6}{6} = {r},
      cell{7}{3} = {r},
      cell{7}{4} = {r},
      cell{7}{5} = {r},
      cell{7}{6} = {r},
      cell{8}{3} = {r},
      cell{8}{4} = {r},
      cell{8}{5} = {r},
      cell{8}{6} = {r},
      hline{1,3,9} = {-}{},
      hline{2} = {3-6}{Gray},
      hline{6} = {-}{Gray},
    }
    \textbf{Dataset}              & \textbf{Label} & \textbf{Text Length (Characters)} &         &         &         \\
                                  &                & Mean                              & Std Dev & Maximum & Minimum \\
    Wikipedia Personal
      Attacks  & All            & 402                               & 734     & 10000   & 1       \\
                                  & Positive       & 341                               & 938     & 10000   & 5       \\
                                  & Negative       & 410                               & 702     & 10000   & 1       \\
    IMDb Movie Review
      Sentiment & All            & 1297                              & 980     & 14106   & 32      \\
                                  & Positive       & 1303                              & 1014    & 13593   & 65      \\
                                  & Negative       & 1271                              & 928     & 8698    & 32      
    \end{tblr}
    \caption{Statistics of text length across labels for the two datasets.}
    \label{tab:text_length}
\end{table}

For both datasets, we use the existing train splits (breakdowns visible in Table \ref{tab:data_counts}) to create seven training sets of increasing size ([16, 32, 64, 128, 256, 512, 1024]), in order to assess performance at different labelling budgets. For the Wikipedia Personal Attacks dataset, we create two versions of these seven training sets: one with the original balance of labels (11.7\% positive, 88.3\% negative), and one with an even balance (50\% positive, 50\% negative), in order to provide insight into the impact of balanced vs. unbalanced data on technique performance. As the original test data is large and performing inference on such large data is time-consuming, we create a final test set for each dataset from a 10\% stratified random sample (preserving the original class balance) of the original data, resulting in a final test set of 2,316 entries for Wikipedia Personal Attacks, and 1,250 entries for IMDb Movie Review Sentiment.

\begin{table}[ht]
    \centering
    \begin{adjustbox}{width=1\textwidth}
    \footnotesize 
    \begin{tblr}{
      row{2} = {c},
      cell{1}{1} = {r=2}{},
      cell{1}{2} = {r=2}{},
      cell{1}{3} = {c=3}{c},
      cell{3}{1} = {r=5}{},
      cell{3}{3} = {r},
      cell{3}{4} = {r},
      cell{3}{5} = {r},
      cell{4}{3} = {r},
      cell{4}{4} = {r},
      cell{4}{5} = {r},
      cell{5}{3} = {r},
      cell{5}{4} = {r},
      cell{5}{5} = {r},
      cell{6}{3} = {r},
      cell{6}{4} = {r},
      cell{6}{5} = {r},
      cell{8}{1} = {r=5}{},
      cell{7}{3} = {r},
      cell{7}{4} = {r},
      cell{7}{5} = {r},
      cell{8}{3} = {r},
      cell{8}{4} = {r},
      cell{8}{5} = {r},
      cell{9}{3} = {r},
      cell{9}{4} = {r},
      cell{9}{5} = {r},
      cell{10}{3} = {r},
      cell{10}{4} = {r},
      cell{10}{5} = {r},
      cell{11}{3} = {r},
      cell{11}{4} = {r},
      cell{11}{5} = {r},
      cell{12}{3} = {r},
      cell{12}{4} = {r},
      cell{12}{5} = {r},
      hline{1,3,13} = {-}{},
      hline{2} = {3-5}{Gray},
      hline{8} = {-}{Gray},
    }
    \textbf{Dataset} & \textbf{Split} & \textbf{Label} & & \\
      & & \textit{All} & \textit{Positive} & \textit{Negative} \\
    Wikipedia Personal Attacks & All & \textbf{115,864} & \textbf{13,590  (11.73\%)}   & \textbf{102,274  (88.27\%)} \\
     & Train & \textbf{69,526} & 8,079  (11.62\%) & 61,447  (88.38\%) \\
     & Test & \textbf{23,178} & 2,756  (11.89\%) & 20,422  (88.11\%) \\
     & Development & \textbf{23,160} & 2,755 
      (11.90\%) & 20,405  (88.10\%) \\
     & Exploration (WS) & \textbf{100} & 12  (12\%) & 88  (88\%) \\
    IMDb Movie Review Sentiment & All & \textbf{50,000} & \textbf{25,000  (50\%)} & \textbf{25,000  (50\%)} \\
     & Train & \textbf{25,000} & 12,500  (50\%) & 12,500  (50\%) \\
     & Test & \textbf{12,500} & 6,250  (50\%) & 6,250  (50\%) \\
     & Development & \textbf{12,500} & 6,250  (50\%) & 6,250  (50\%) \\
     & Exploration (WS) & \textbf{100} & 92  (92\%) & 8  (8\%) \\
    \end{tblr}
    \end{adjustbox}
    \caption{Counts of dataset entries across splits and labels. The exploration data set is exclusive for the Weak Supervision Technique.}
    \label{tab:data_counts}
\end{table}

An extra split is added exclusively for the weak supervision technique. This labelled \textit{exploration} test is intended to test and fine-tune labelling functions then used in the training process. In our case, we take 100 data points for exploration of each task. Details of these sets are detailed in Table \ref{tab:data_counts} and these additional points should be considered in the labelling budget. The exploration set is drawn from the training set, making sure that the data points in this set never overlap with the 7 training sets mentioned in the previous paragraph.

\subsection{Model Selection and Training Details}

For the conventional baseline models, we develop a Multinomial Na\"ive Bayes classifier (typically used for text classification - e.g., to represent how many times a word appears in a document) built based on Bayes's theorem \citep{10.1007/978-3-540-30549-1_43}.\footnote{We use the package developed by scikit-learn: \url{https://scikit-learn.org/stable/modules/generated/sklearn.naive_bayes.MultinomialNB.html}}. A deeper analysis on why we choose the Multinomial Na\"ive Bayes and not other options like the presented in \cite{rennie_tackling_nodate} and \cite{fatourechi2008comparison} can be read in the Appendix \ref{ssec:naive_bayes} of the online supplement. For the logistic regression classifier, we construct a logistic regression with an L2 penalty parameter cross-validated to maximise the used F1 macro score\footnote{We use the package developed by scikit-learn: \url{https://scikit-learn.org/stable/modules/generated/sklearn.linear_model.LogisticRegression.html}}. These methods are selected because they can be easily developed and speedily trained without using GPUs. We use TF-IDF to convert data into vectors/features.

For weak supervision, we use the \texttt{LabelModel} presented in \citet{ratner_training_2019, ratner_data_2016} and implemented in the \href{https://snorkel.readthedocs.io/en/v0.9.3/packages/_autosummary/labeling/snorkel.labeling.LabelModel.html}{Snorkel package} \citep{ratner_snorkel_2017}. This model takes the label matrix obtained by applying the labelling functions to the training data set and uses a Bayesian probabilistic model and a matrix completion algorithm to obtain a set of weights. These weights are trained to minimise the error between the golden labels from the training data set and the labels obtained by the weighted sum of labelling functions over the data set. Once the model is trained, it is applied to the test set.   

For both transfer learning and prompt engineering, there is a need to choose a pre-existing model. For both techniques, we chose pre-trained \texttt{Distilbert-base-cased} (a distilled version of BERT, \citet{devlin-etal-2019-bert}) as a base model since it retains 97\% of BERT performance on several natural language tasks while being 40\% smaller and 60\% faster \citep{sanh2019distilbert}. Such characteristics are in line with our goals of developing applications with economic budgets requiring less training time, computation and memory. In addition to the base model, we conducted additional sets of experiments using \texttt{DeBERTa-v3} \citep{2021arXiv211109543H} and \texttt{GPT-2} \citep{radford2019language} for transfer learning and prompt engineering, respectively. DeBERTa-v3 is a DeBERTa model with improvements based on replaced token detection and a new weight-sharing method. GPT-2 is an autoregressive transformer-based language model trained on a dataset of 8 million web pages using a next-word prediction objective. While DeBERTa-v3 and GPT-2 have larger model sizes and slightly different model structures and training procedures, both have shown superior performance on downstream tasks \citep{10045_127431}. By including a variety of starting points, we can study the impacts of using different base models on our classification tasks.

For our zero-shot learning experiments, meanwhile, we make use of GPT-3, 3.5 and 4, as provided through the \href{https://platform.openai.com/}{\texttt{OpenAI API}}. We make use of these because, at the time of writing, they are widely regarded as amongst the state of the art in terms of large language models and thus likely to offer the best performance for the task at hand. As they are available through an API, they can also be used by anyone with an internet connection, hence size and computational costs are much less of a consideration (we will comment on the actual cost of accessing the API below). 

\subsection{Hyperparameter selection and training details}
 
We fine-tuned all models selected for transfer learning and prompt engineering for three epochs with the default set of hyperparameters: a learning rate of 1e-3, a maximum text length of 512 tokens and a batch size of eight. To facilitate reproducibility, we conducted three trials of each experimental configuration using seeds [1, 2, 3], with only the seed varied. We use the transformers library \citep{wolf-etal-2020-transformers} for transfer learning and the OpenPrompt library \citep{ding-etal-2022-openprompt} for prompt engineering. Our experiments were performed on a single NVIDIA K80 GPU with 56 GiB memory.

Specific to prompt engineering, the training data is structured as $[text] \; Prompt \; [class]$. As discussed above, finding the best prompt is not trivial and requires trial and error (though previous work shows that prompt choice is not a dominant factor in model performance compared with hyper-parameters such as random seeds, \citet{le-scao-rush-2021-many}). To try and mitigate the possibility that our results are based on the specifics of an arbitrary prompt template, we test three prompt variations for each task. The prompts are varied in pattern and language style while being valid descriptions without changing the semantics, representing real-world user scenarios. For abuse classification, the prompts are, `Is this text abusive?', `Does this text contain abuse?', and `It was? Abusive or Not Abusive'. For movie sentiment classification task, the prompts are, `Is this text negative?', `Does this text contain negative sentiment?', and `It was? Negative or Not Negative'. Having a variety of prompts allows us to show how performance might vary between different reasonable prompts. 

OpenAI's API also has a number of hyperparameters that may affect the output provided by the model \citep{OpenAI}. For example, previous work has evaluated generated text on varying levels of temperature, top p, and max tokens \citep{bommarito2022gpt,rehana2023evaluation}. Similar to \citet{rehana2023evaluation}, we set temperature, which affects how deterministic the output text is, to 0.1. While temperature is above zero the results will not be purely deterministic, however performance has been shown to improve empirically. While each of our prompts attempts to elicit a single-word response, we set the maximum number of returned tokens to 20 in order to examine why non-response occurs within the model. We additionally provide a number of stop tokens, such as a period and a comma, to limit as much as possible the model's tendency to provide more generated text beyond the single-word response we ask of it. 
\section{Results}
\label{sec:results}

In this section, we present the results of our experiments, to illustrate these techniques in action. Although our experiments made use of multiple models and configurations, in order to simplify presentation we display just a sample of these. In particular, for transfer learning we show results generated using \texttt{DistilBERT}, and for prompt engineering we also look at \texttt{DistilBERT} and look at only one fixed prompt (`Does this text contain abuse?' for abuse classification and `Does this text contain negative sentiment?' for movie sentiment classification). For Na\"ive Bayes, logistic regression and weak supervision, we present the full results. Other base models and prompts provided similar results.

\begin{figure}[ht]
    \centering
    \includegraphics[width=\textwidth]{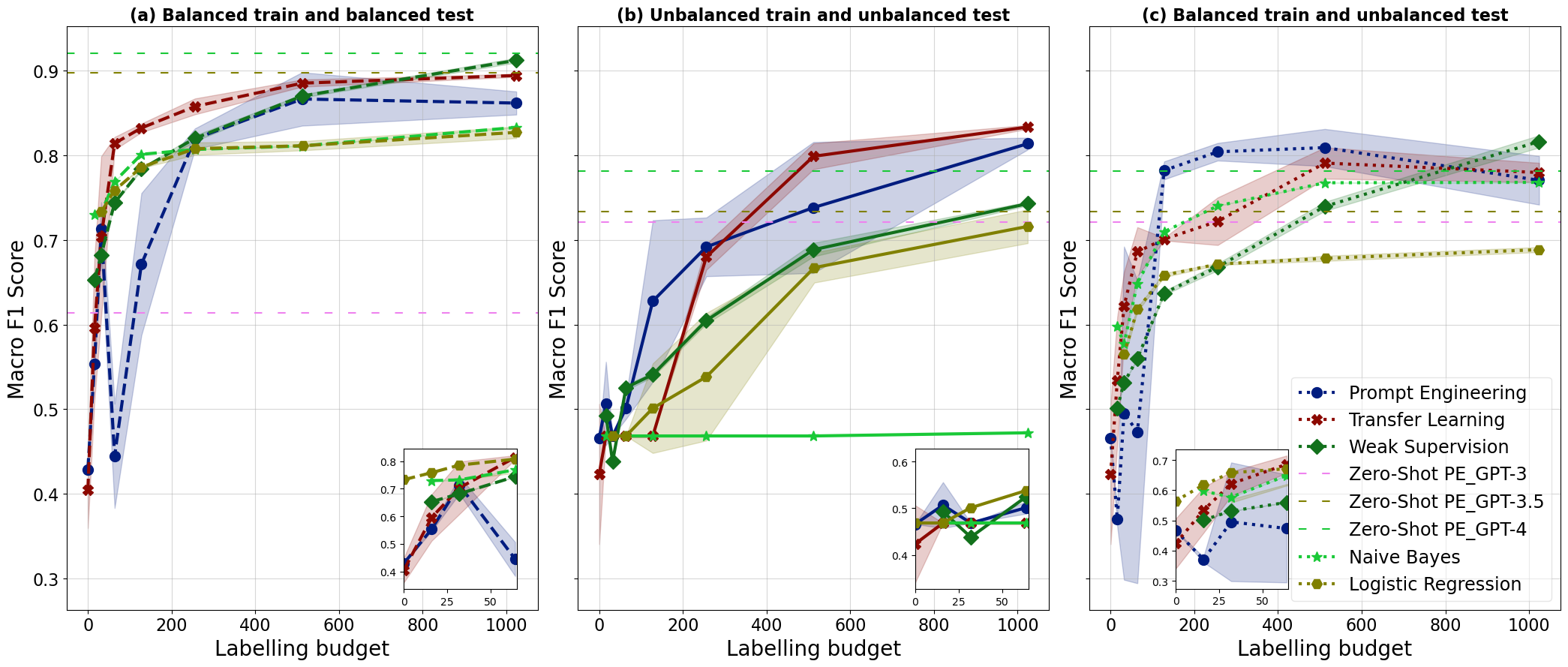}
    \captionsetup{justification=centering,margin=2cm}\caption{\label{fig:binary_abuse_method_comparison_lc}Learning curves of binary abuse classification task.}
\end{figure}

We begin by investigating our binary abuse classification task, shown in Figure \ref{fig:binary_abuse_method_comparison_lc}. This figure has several panels, reflecting whether the underlying data is balanced or not. We can see that in the simple case of underlying balanced data (Figure \ref{fig:binary_abuse_method_comparison_lc}, Panel (a)), which will hence generate both balanced training datasets and test datasets), there is actually little difference between any of the techniques in question. The `conventional' machine learning approaches achieve a high performance (Macro F1 of around 0.8) with only around 150 examples. While it is eventually outperformed by prompt engineering, transfer learning and weak supervision, this is only after the labelling budget crosses the 250 mark\footnote{when interpreting the labelling budgets, we should also bear in mind that, in addition to the training budget, we need to consider the 100 data points used in the exploration data set for the weak supervision labelling functions}, and even after 1,000 examples have been used the difference is not enormous. Zero-shot learning with GPT-3.5 and GPT-4 offers an impressive Macro F1 (around 0.9) without any training data. 

However, we know that the prevalence of abusive text is generally lower than the presence of non-abusive text. Hence a more `realistic' scenario is that both training and test data are unbalanced (Figure \ref{fig:binary_abuse_method_comparison_lc}, Panel (b)). Here we can see that conventional machine learning performs worse than its `cheaper' counterparts, with a Macro F1 score of less than 0.5 for all levels of the training budget investigated for the Na\"ive Bayes classifier, and with high standard deviation and a lower Macro F1 score than the rest for the case of the logistic regression. In the unbalanced case, where abuse is around 10\% of the total dataset, then even with a relatively large labelling budget only a small amount of cases from the positive class will be seen. The difference between the other techniques in particular is notable, with transfer learning able to achieve the highest overall performance (though the Macro F1 of 0.8 from GPT-4 is again notable, considering that no training data is required to achieve it). 

Of course, in the case of unbalanced data, one common workaround is to artificially balance the training dataset (though of course preserving the real underlying distributions of data in the test data). This can improve model performance on a limited training budget (though achieving training balanced data still requires a higher labelling effort, or trying to artificially inflate the likelihood of labelling positive data through the use of keywords). Results from this approach are presented in Figure \ref{fig:binary_abuse_method_comparison_lc}, Panel (c). We can see that model performance is improved in all cases, reaching the range of 0.7-0.8 after a few hundred examples (the zero-shot approaches all fall within this range as well), except for the logistic regression results. Weak supervision is actually the best-performing technique in this scenario (as the test data is the same, the results for zero-shot prompt engineering are the same in Panels (b) and (c)). 

\begin{figure}[ht]
    \centering
    \includegraphics[width=\textwidth]{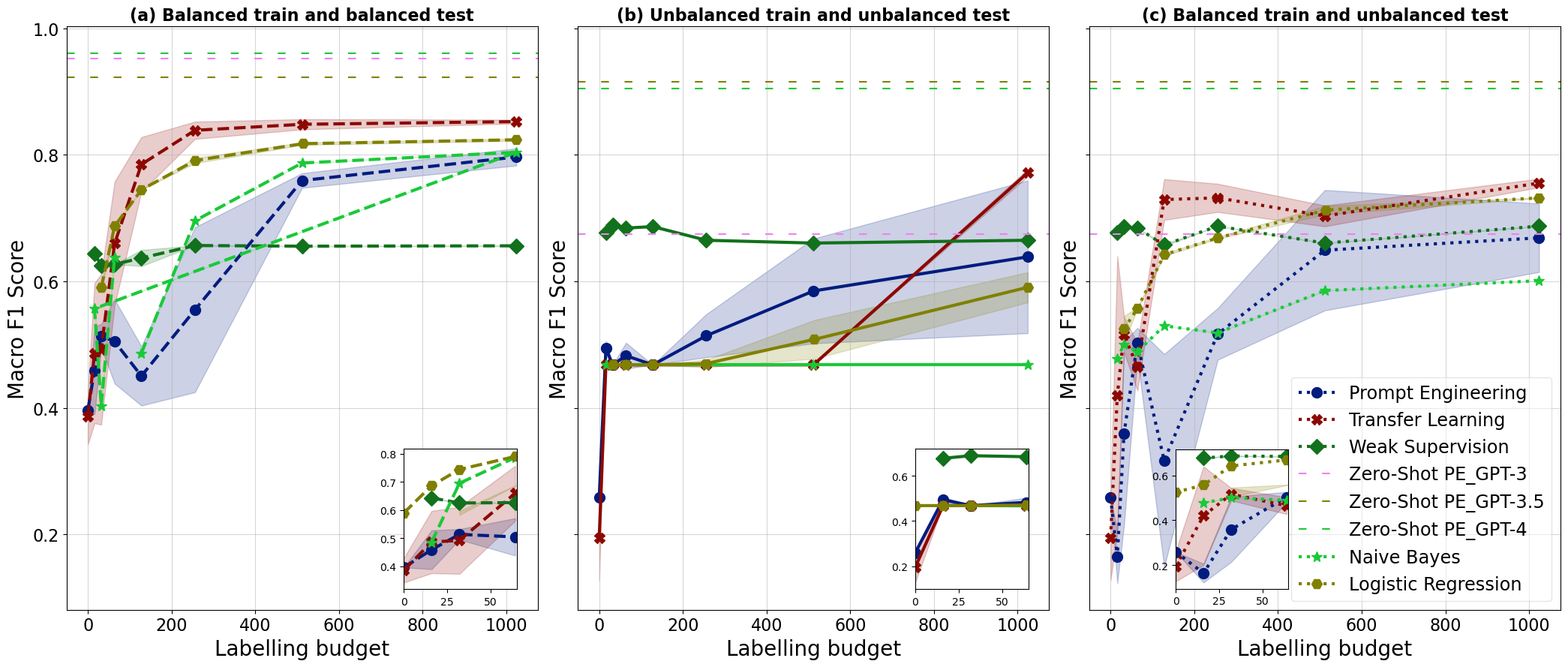}    \caption{\label{fig:binary_movie_method_comparison_lc}Learning curves of binary movie sentiment classification task.}
\end{figure}

We will now move on to address our binary movie sentiment classification task, shown in Figure \ref{fig:binary_movie_method_comparison_lc}. The format of this figure is the same, with panels addressing classifier performance over a variety of labelling budgets for balanced data, unbalanced data, and a balanced training set with unbalanced test data. Again, these three setups were chosen as representative of common real-world research scenarios. The overall results from this task are similar to the abuse classification task. For balanced data, we can see that all of the methods perform in a similar fashion. 
Our conventional Na\"ive Bayes approach achieves a Macro F1 below 0.7 with around 100 labelled examples. Prompt engineering, transfer learning and weak supervision are able to outperform it, especially as the labelling budget increases, eventually arriving at scores of around 0.9. However, in this case the logistic regression achieves as good results as transfer learning. Again, zero-shot learning performs well, with performance above 0.9 for GPT-4. In the unbalanced case, the performance of all models is lower, though transfer learning and prompt engineering can achieve a Macro F1 of 0.8 with around 1,000 examples, again common with GPT-4. Our Na\"ive Bayes model performs very poorly in this context. Finally, when training data is balanced but test data remains unbalanced, more of the approaches can achieve a Macro F1 of around 0.8, though we would recall again that the cost of generating a balanced training dataset from unbalanced data can be quite high. \\

The IMDB movie sentiment database \citep{maas_learning_2011} has been, as mentioned in the Methods section \ref{ssec:data}, extensively studied in the literature. As such, a natural worry arises that the database has been used in the training datasets of GPT-3.5 and GPT-4, thus contaminating the zero-shot testing described above. Given that we are unable to check if indeed the database was used to train the GPT models, we collected an analogous validation of movie reviews from another review aggregator, TMDB\footnote{https://www.themoviedb.org/}. The included reviews were publicly published from October 2021, past the training date cut-off for the GPT-3.5. As in \citet{maas_learning_2011}, we label as positive those reviews with a score higher than 6 out of 10, and we label negative those reviews leaving a score lower than 5, omitting neutral reviews and including no more than 30 reviews from a single movie. Our final database comprises 855 reviews, with a distribution of 73.3\% of positive reviews and 26.7\% of negative reviews. When performing a zero-shot test on the TMDB dataset, we obtain Macro F1 scores between 0.87 and 0.94 for both GPT-3.5 and GPT-4, using the three prompts as the other tasks. GPT-3 has, on the other hand, different performances depending on the prompt, with Macro F1 scores of 0.67, 0.85 and 0.11\footnote{he three prompts used are shown in Table \ref{tab:results_tmdb}. The first two (and best) results correspond to simple questions without any framing to the LLM. On the contrary, the third question giving a poor result (0.11 F1 Macro) corresponds to a prompt with a framing of the problem we are facing. This would suggest that, contrary to newer models, GPT-3 does not react well to framing.}. The results are in accordance with the results presented in Figure \ref{fig:binary_movie_method_comparison_lc}, thus validating them for our use of the three versions of GPT. In other words, we are confident the high performance of GPT in these classification tasks is not solely a result of GPT's training data already including the IMDB dataset.  \\

\begin{figure}[ht]
    \centering
        \includegraphics[width=\textwidth]{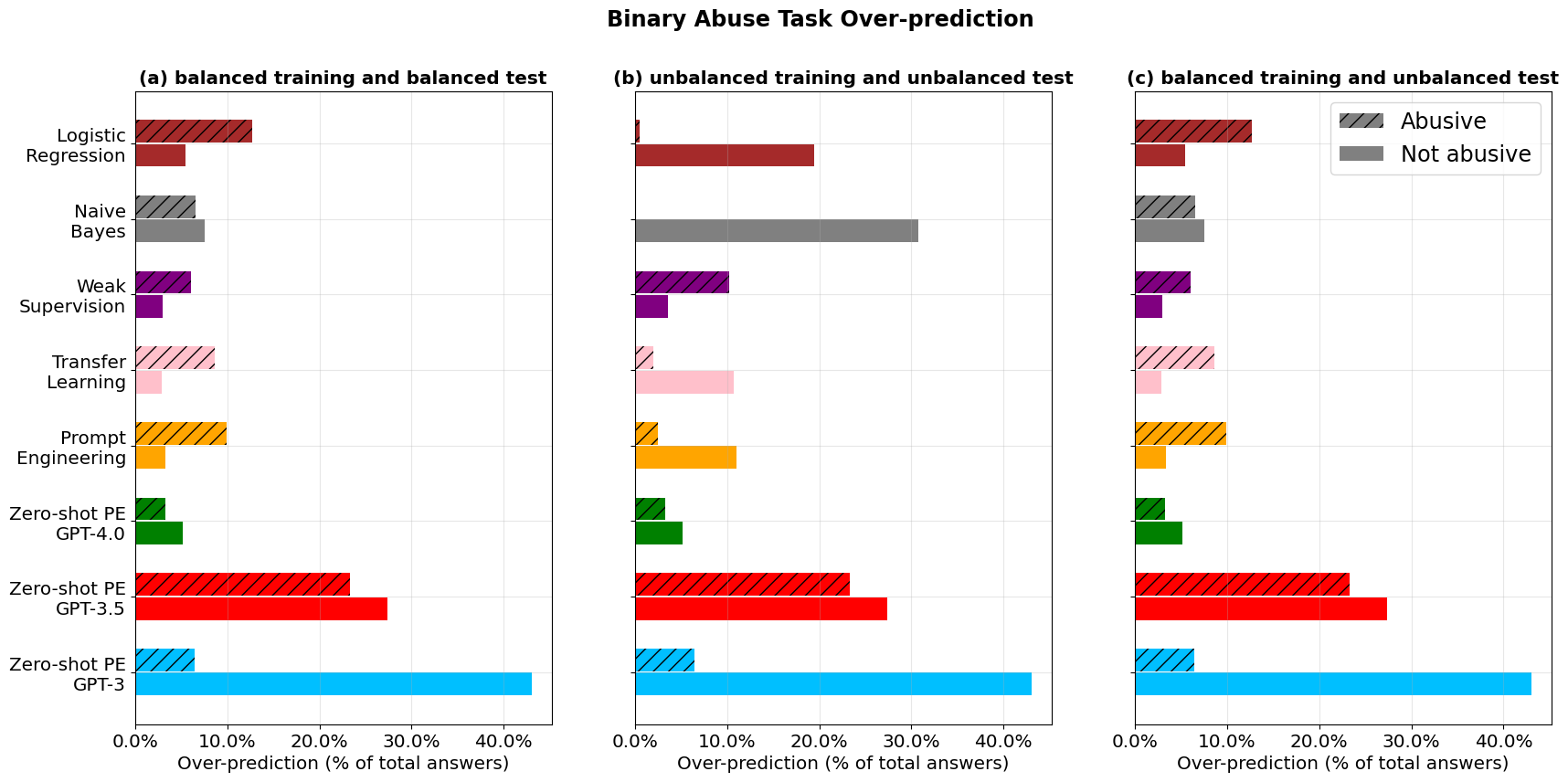}
    \caption{Over-prediction bar plots for every technique used when performing the binary abuse task. For non-zero shot techniques, we use the highest labelling budget $n=1024$.}
    \label{fig:over-predictions_binary_abuse}
\end{figure}

\begin{figure}
    \centering
    \includegraphics[width=\linewidth]{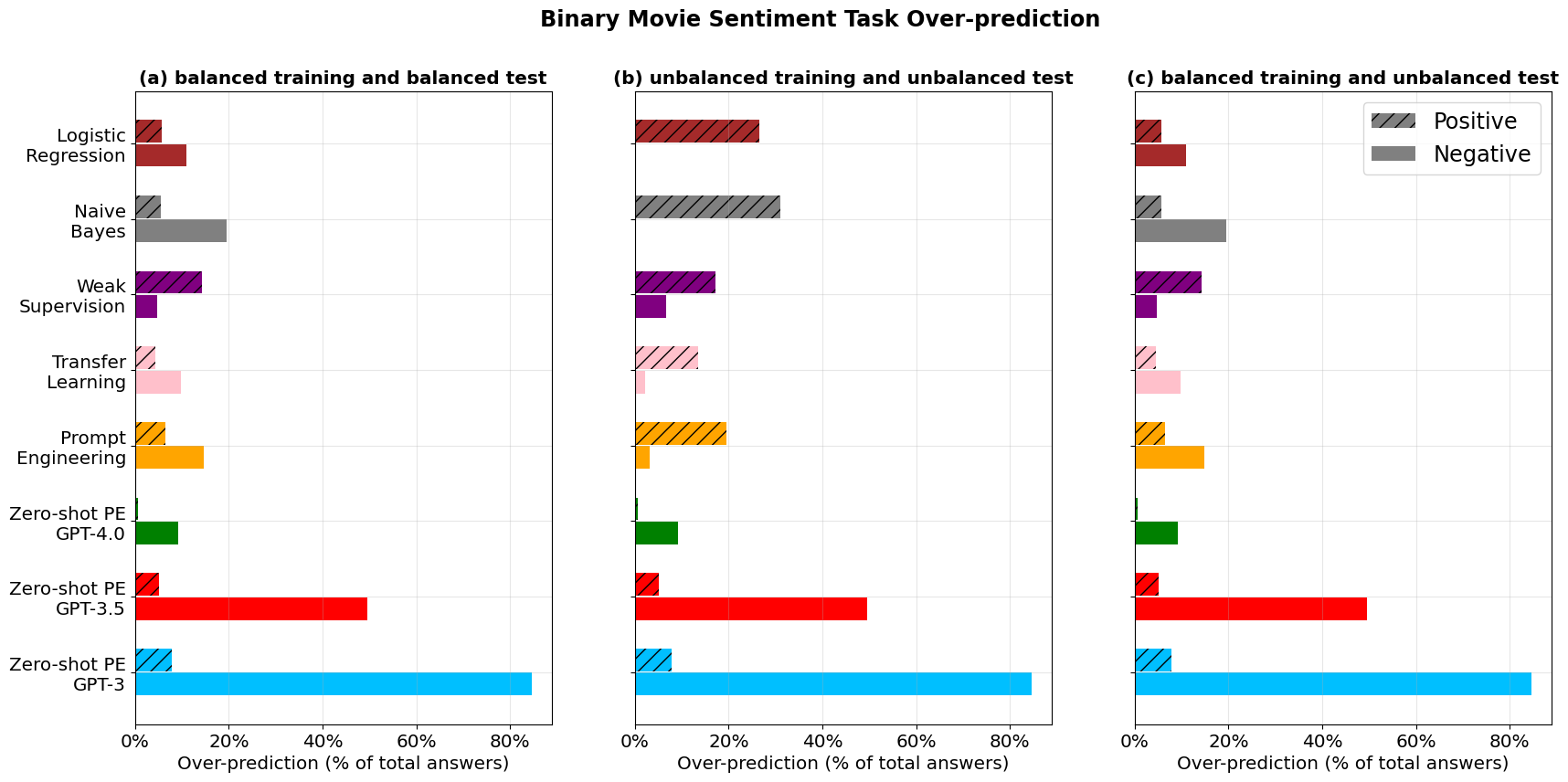}
    \caption{Over-prediction bar plots for every technique used when performing the binary movie sentiment task. For non-zero shot techniques, we use the highest labelling budget $n=1024$.}
    \label{fig:over_predictions_binary_movie_sentiment}
\end{figure}

The bias analysis performed for both tasks and all the techniques here is inspired by the work of \cite{ashwin_using_2023}, where the authors examine the biases that open and closed LLMs have when annotating refugees' interviews. We compute the over-prediction of each class in our studied tasks. For the non-zero-shot prompt engineering techniques, we compute the over-prediction at the highest labelling budget ($n=1024$). Given the good results of the zero-shot prompt engineering shown in Figures \ref{fig:binary_abuse_method_comparison_lc} and \ref{fig:binary_movie_method_comparison_lc}, we compare the biases of the different techniques at their best performances to give the reader an extra dimension with which to compare the studied techniques other than their performance scores. Results are shown in Figures \ref{fig:over-predictions_binary_abuse} and \ref{fig:over_predictions_binary_movie_sentiment}.

While the over-prediction percentages vary between each task --particularly for the zero-shot PE with GPT-3.0--, results remain qualitatively similar allowing us to draw general conclusions. 
The GPT-3 technique has the highest over-prediction for both tasks, not detecting abusive language (binary abuse task) or positive language (binary movie sentiment task), while having a relatively low percentage (<10\%) of over-prediction for the opposite classes. 
The highest over-prediction for both classes is obtained by GPT-3.5 (ChatGPT) in the case of the binary abuse task, while presenting a similar behaviour as GPT-3 for the binary movie sentiment. 
Finally, GPT-4.0 presents the lowest over-predictions for both tasks. This is interesting because, while the F1-macro scores between GPT-3.5 and GPT-4.0 are relatively similar in Figure \ref{fig:binary_abuse_method_comparison_lc} and \ref{fig:binary_movie_method_comparison_lc}, the over-prediction analysis gives completely different behaviours for both models. 

The classical machine learning techniques (Na\"ive Bayes and logistic regression) over-predict one class when there is an unbalanced training dataset, making them sensitive to class unbalance.
Finally, the Weak Supervision, Transfer Learning and Prompt Engineering techniques show similar behaviours, with over-predictions lower than 20\% for both cases and an inversion of the most over-predicted class for balanced and unbalanced training datasets. \\

\begin{figure}[ht]
    \centering
    \captionsetup{justification=centering,margin=2cm}
    \includegraphics[width=0.75\textwidth]{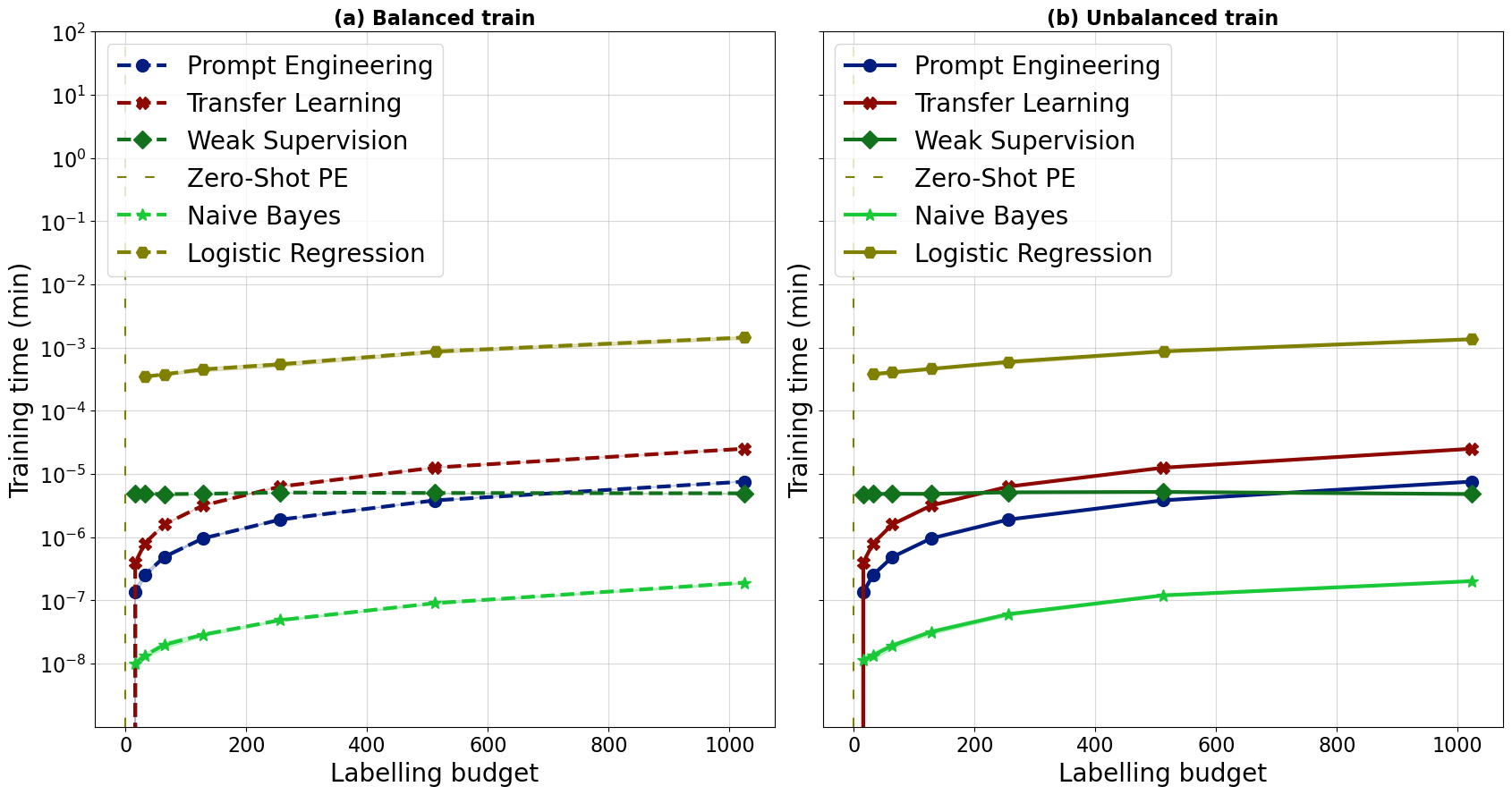}    \caption{\label{fig:binary_abuse_method_comparison_times}Training time of binary abuse classification task.}
\end{figure}

\begin{figure}[ht!]
    \centering
    \captionsetup{justification=centering,margin=1cm}
        \includegraphics[width=0.75\textwidth]{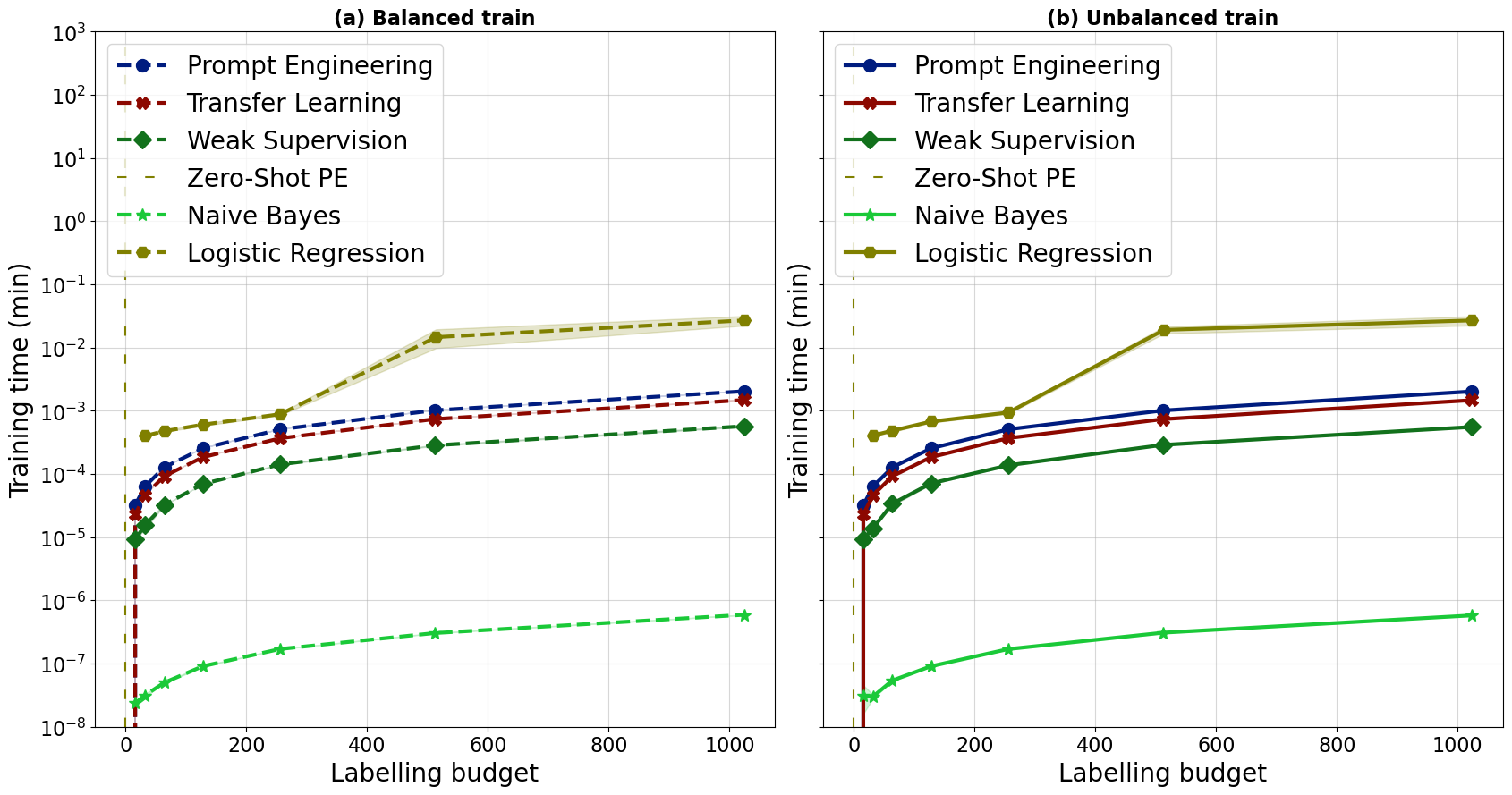}        \caption{\label{fig:binary_movie_sentiment_method_comparison_time}Training time of binary movie sentiment classification task.}
\end{figure}

It is also worth evaluating the amount of training time required for each of the techniques, something which has knock-on consequences for the financial costs involved if making use of a cloud platform. Overall, the training time is short for all three techniques and two datasets (see Figures \ref{fig:binary_abuse_method_comparison_times} and \ref{fig:binary_movie_sentiment_method_comparison_time}). As the labelling budget grows, the training time also grows linearly except for weak supervision. For weak supervision, increasing labelling examples does not seem to affect training time when trained on Wikipedia personal attacks dataset. The training time is generally around $10^{-5}$ minute regardless of labelling budgets and slightly decreases when using the maximum labelling budget in our experiments ($n$=1024). 
Training times in this case are a sum of two elements: the time that the set of labelling functions takes to label data points and the time that the probabilistic \texttt{LabelModel} \citep{ratner_training_2019} takes to compute the weights and probabilities associated to the labelling functions. The different behaviours seen in Figure \ref{fig:binary_abuse_method_comparison_times} and Figure \ref{fig:binary_movie_sentiment_method_comparison_time} depend on the nature of the labelling functions used. In the case of the binary movie sentiment analysis, we add two \texttt{regex} LFs, thus observing the linear increase of time as they require more time to look for matches as the number of data points increases. In the case of the binary abuse experiment, the labelling functions used are dictionary-based or transforming annotators' knowledge into a labelling function. Given that directory-based functions (looking for words in a text) are not as computationally expensive as regular expression functions, given that the maximum length of the text is fixed, we obtain a stable training time of 1 minute. 

It is also important to note how the conventional machine learning methods bound the training times of all the others. While Na\"ive Bayes training time is the lowest of the bunch, the logistic regression is the highest, with a difference between the two of almost 5 orders of magnitude for both tasks. However, the training times for all techniques are still relatively low, below 0.1 minutes.

While Zero-shot Prompt Engineering does not require any training time, it is worth briefly commenting on the costs that could be incurred through the API to perform classification tasks (these are correct at the time of writing). Costs through such an API are per `token': a small piece of text that might be equivalent to a word, a part of a word or a piece of punctuation (as a rule of thumb, a token is often around four characters long). To perform a classification using an LLM there will be costs for both input tokens (the chunk of text to be classified) and output tokens (the response indicating the classification). Of course, the response is usually just one or two tokens long, meaning the length of the input text is the main factor determining API costs. At the time of writing, GPT-3 had a cost of £0.0006 per 1,000 tokens, GPT-3.5 a cost of £0.0013 per 1,000 tokens and GPT-4 a cost of £0.0025 per 1,000 tokens. The price for a project will, obviously, vary according to the model chosen and the type of input text but, to give an example, the movie reviews were on average around 325 tokens each, meaning that 10,000 movie reviews could be classified on GPT-4 for around £8. Hence, API costs are only likely to be a significant concern for relatively large pieces of text where there are millions of observations to be classified. 
In the case in which we would like to train the newer OpenAI models (GPT-3.5, GPT-4.0) just as we did with GPT-2.0 for both datasets, we would also need to consider the API costs of the training data sets into account. However, given the good results that the zero-shot learning experiments give, we did not consider these scenarios which would represent an extra cost and thus would go against the spirit of this study.

\section{Conclusions}

In this paper, we have sought to give an overview of three methods for automatic content analysis that are `cheaper' than the conventional approach to machine learning, in the sense that they seek to limit the amount of labelled data that is required as a basis for training a model. In this final section, we will draw some conclusions about the significance of our experimental results, and future directions for automatic content analysis in the field. 

First, it was striking how all of the techniques we reviewed achieved strong performance in most of the tasks we reviewed, on small labelling budgets. In most cases, a few hundred data points were enough to achieve a Macro F1 score of above 0.7. It should be noted however that the conventional baseline we addressed also performed well for many tasks on similar labelling budgets, provided the data were balanced (though we should emphasise that unbalanced data scenarios are very common in automatic content analysis). 

Second, perhaps most of all, it was striking how zero-shot prompt engineering performed well across all of the tasks we reviewed, comfortably outperforming any of the other methods on the movie review task and offering equivalent performance on the abuse classification task. 
However, even if their performances are high, over-predictions and thus systematic biases can also be present. An analysis of these elements are needed no matter the technique used to avoid arriving at incorrect conclusions.
Considering these techniques require no labelled training data (though still require test data) and are very easy to implement, they appear to offer a very promising and flexible solution to automatic content analysis in the future.
Their performance and biases on a wider variety of social science concepts need to be tested; and it should also be borne in mind that there is a financial cost to using them (though not one that will be significant unless millions of lengthy text documents are being labelled). Furthermore, larger models than the current generation of LLMs are already in development, and it is reasonable to expect both performance gains and price decreases as they emerge (in all of our experiments, the increase in size from GPT-3 to GPT-4 corresponded with an increase in performance and a decrease in over-prediction). As this next generation emerges, it may be that other methods of automatic content analysis will start to become redundant, though this proposition would need to be rigorously tested. This proposition would be of particular interest as open- and closed-source LLMs can present serious biases that could skew their uses \citep{spirlingWhy2023, ashwin_using_2023}.

Lastly, despite the impressive performance of zero-shot prompt engineering using LLMs, it is important to assess LLMs across various types of tasks to understand their capacity and limitations. For instance, our results show that model performance and over-predictions can vary across seeds or the prompts tested. This instability further highlights the importance of robustness and reproducibility in LLMs. Furthermore, LLMs can be biased towards certain demographics or identities when, for example, asked to perform curriculum screening \citep{dastin2022amazon, veldanda2023emily}, or annotating refugees' interviews \citep{ashwin_using_2023}. In this paper, we solely consider binary classification tasks as our intention is to keep results as relatable as possible so that the reader can understand the concepts and bring this knowledge to their classification tasks, as complex as they could be. While an open question remains on how LLMs perform on more complicated tasks that involve language understanding such as logical reasoning and inference,
we contribute by giving social scientists different strategies to tackle challenges where data scarcity and bias might be present like analysing interview data from under-represented groups \citep{deterding_flexible_2021,  Small_qualitative_2022, bonikowski_ends_2022, ashwin_using_2023, atari_which_2023} or text analysis and labelling of historical archives \citep{jaillant_archives_2022, marciano_archival_2018, rahal_rise_2024}.
\bigskip


\paragraph*{Acknowledgments.} 
This work was supported by the Public Policy Programme of The Alan Turing Institute under the EPSRC grant EP/N510129/1.
This work was supported by the Ecosystem Leadership Award under the EPSRC Grant EP/X03870X/1 and The Alan Turing Institute. Funded by the UK Government.

\paragraph*{Data Availability Statement.} Our code and experimental results are available here at \url{https://github.com/Turing-Online-Safety-Codebase/cheap_learning}, allowing users to understand, verify and replicate findings.

\paragraph*{Conflict of Interest Statement.} The authors declared no potential conflicts of interest with respect to the research, authorship, and publication of this article.

\bibliographystyle{plainnat}
\bibliography{main}

\appendix
\section{Appendix} 
\label{sec:appendix}

To keep the clarity and focus of the main text, we include detailed findings, supplementary tables, and additional analyses in the present Appendix.








\subsection{Binary abuse classification task}
\bigskip
\subsubsection{Transfer learning}
\begin{figure}[ht]
    \centering
    \captionsetup{justification=centering}
    \includegraphics[width=\textwidth]{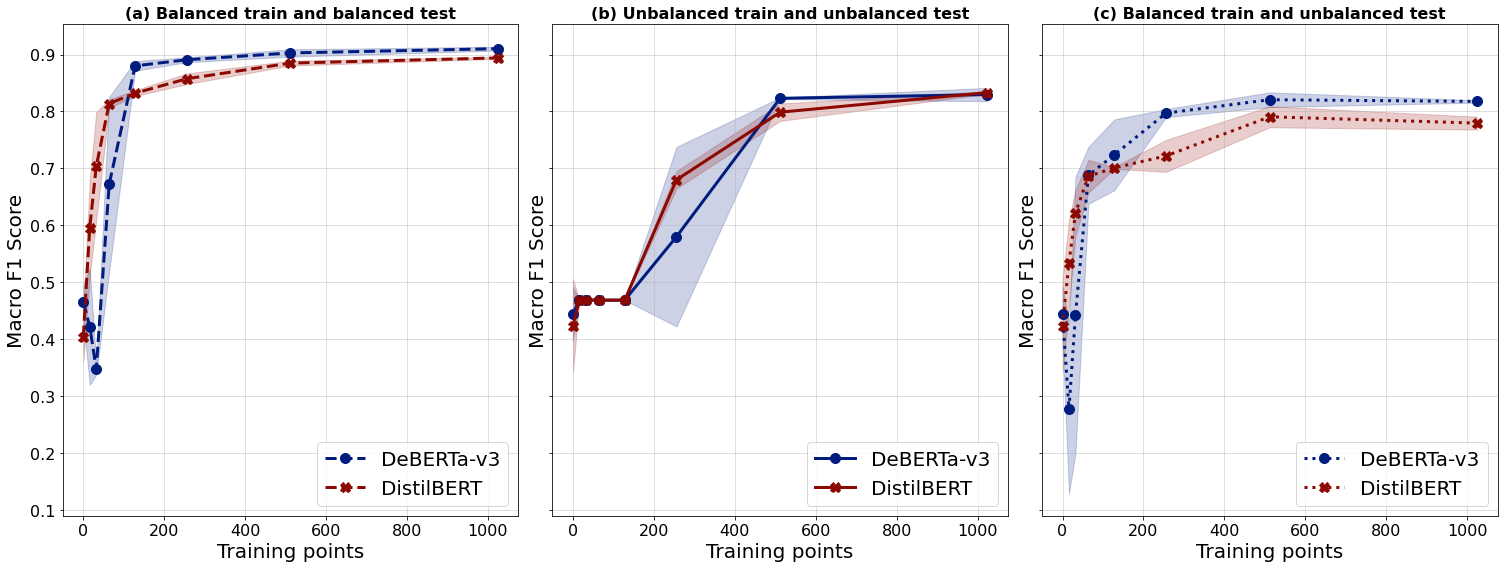}    \caption{\label{fig:binary_abuse_transf_learn_model_comparison_f1}Binary abuse classification results of transfer learning varied by model.}
\end{figure}

In Figure \ref{fig:binary_abuse_transf_learn_model_comparison_f1} we present the learning curves for transfer learning experiments under balanced train and balanced test (\ref{fig:binary_abuse_transf_learn_model_comparison_f1} a), unbalanced train and unbalanced test (\ref{fig:binary_abuse_transf_learn_model_comparison_f1} b), and balanced train and unbalanced test (\ref{fig:binary_abuse_transf_learn_model_comparison_f1} c) using DeBERTa-v3 and DistilBERT. Both models yield similar performance across labelling budgets, while a small fluctuation is observed at 256 training points. With 512 training points models obtain (close to) best results in terms of F1 score, suggesting that 512 can be the amount of data required for reaching optimal performance using transfer learning. When trained on balanced data, both models show a striking jump in performance in the early stage of training with 128 training training points. 

\subsubsection{Prompt engineering}
\begin{figure}[ht!]
    \centering
    \captionsetup{justification=centering}
    \includegraphics[width=\textwidth]{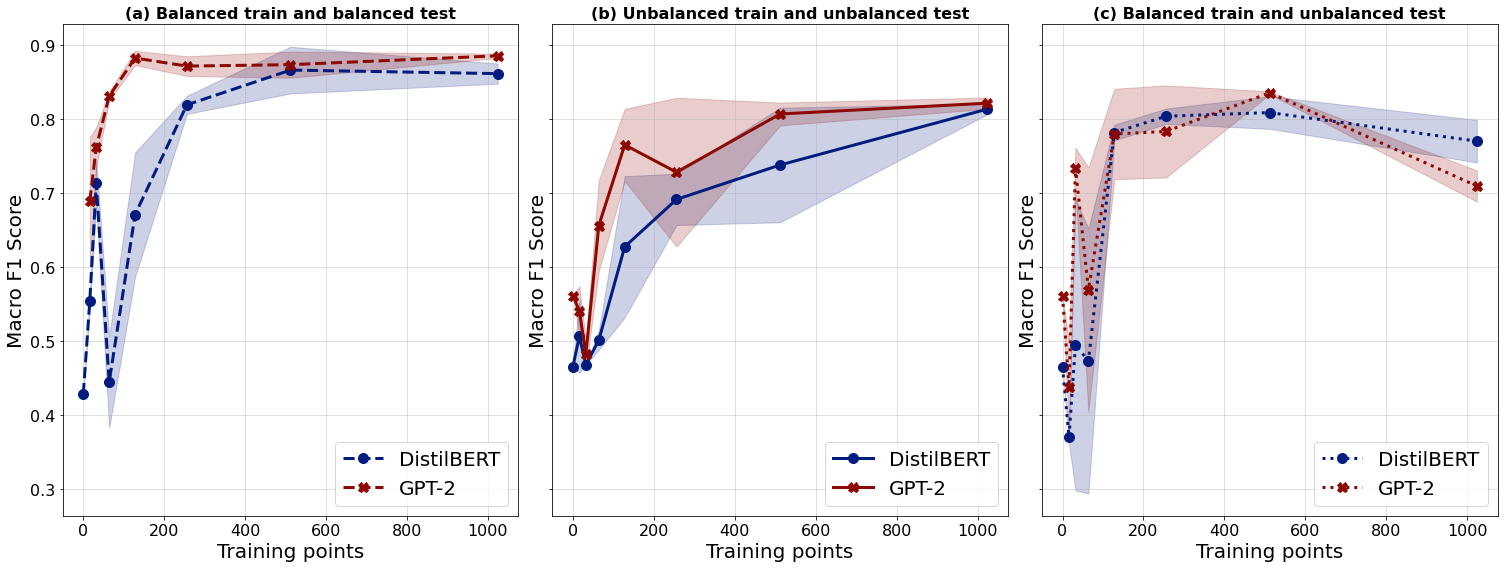}    \caption{\label{fig:binary_abuse_promp_eng_model_comparison_f1}Binary abuse classification results of prompt engineering varied by model.}
\end{figure}

In Figure \ref{fig:binary_abuse_promp_eng_model_comparison_f1} we present the learning curves for prompt engineering experiments under balanced train and balanced test (\ref{fig:binary_abuse_promp_eng_model_comparison_f1} a), unbalanced train and unbalanced test (\ref{fig:binary_abuse_promp_eng_model_comparison_f1} b), and balanced train and unbalanced test (\ref{fig:binary_abuse_promp_eng_model_comparison_f1} c) using GPT-2 and DistilBERT. 
The results are based on the prompt `Does this text contain abuse?'. Generally, GPT-2 obtains better results than DistilBert, which can be due to model structures and pre-training objectives. Similar to transfer learning, models trained under balanced scenarios demonstrate a faster learning curve at distinguishing between classes in comparison with unbalanced settings. Best results are also obtained at 512 training points. In fact, the classification performance of both models drops at the maximum labelling budgets under balanced training scenarios, which shows that more data is not always beneficial for model learning. This emphasizes the advantages of class balance outweigh data size. As reported in previous work \citep{islam2017reproducibility}, high variance across seeds is of particular interest and highlights the need for searching for the best seed.

\begin{figure}[ht!]
    \centering
    \captionsetup{justification=centering}
    \includegraphics[width=\textwidth]{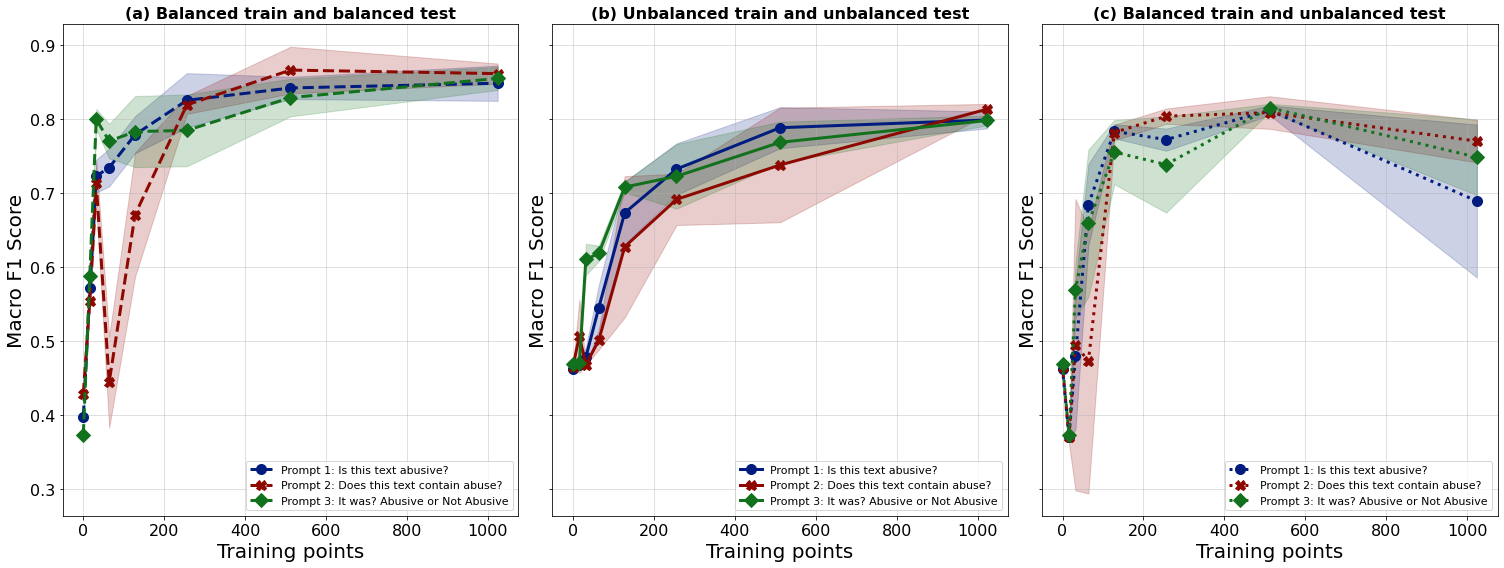}    \caption{\label{fig:binary_abuse_prompt_comparison_f1}Binary abuse classification results of prompt engineering varied by template.}
\end{figure}

To investigate the impact of different prompts on classification, in Figure \ref{fig:binary_abuse_prompt_comparison_f1} we plot the Macro F1 for each of the three prompts under under balanced train and balanced test (\ref{fig:binary_abuse_prompt_comparison_f1} a), unbalanced train and unbalanced test (\ref{fig:binary_abuse_prompt_comparison_f1} b), and balanced train and unbalanced test (\ref{fig:binary_abuse_prompt_comparison_f1} c). 
The results shown are based on Distilbert. Model performance expectedly varies across prompts while demonstrating similar learning curves over labelling budgets. The confidence intervals of the learning curves of each run overlap, which suggests that divergences across seeds can be more important than the potentials of prompt types. This finding is similar to the work done by \citet{le-scao-rush-2021-many}.
Best results achieved are with `Does this text contain abuse?'. The reasons that led to the differences in prompts are unclear. Future work can explore prompt types to improve the classification by changing, for instance, the syntactical, lexical and semantic aspects of prompts.  

\subsection{Binary movie sentiment classification task}

\begin{figure}[ht!]
    \centering
    \captionsetup{justification=centering}
    \includegraphics[width=\textwidth]{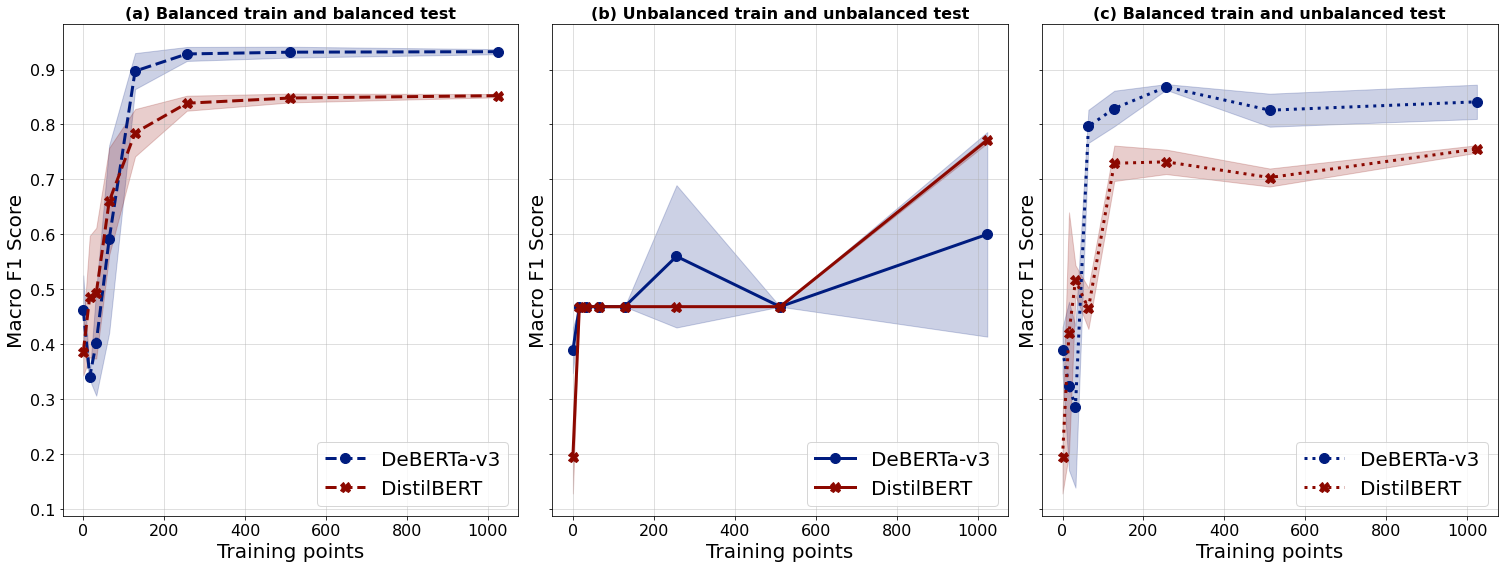}    \caption{\label{fig:binary_movie_sentiment_transfer_learning_f1}Binary movie sentiment classification results of transfer learning varied by model.}
\end{figure}

\begin{figure}[ht!]
    \centering
    \captionsetup{justification=centering}
    \includegraphics[width=\textwidth]{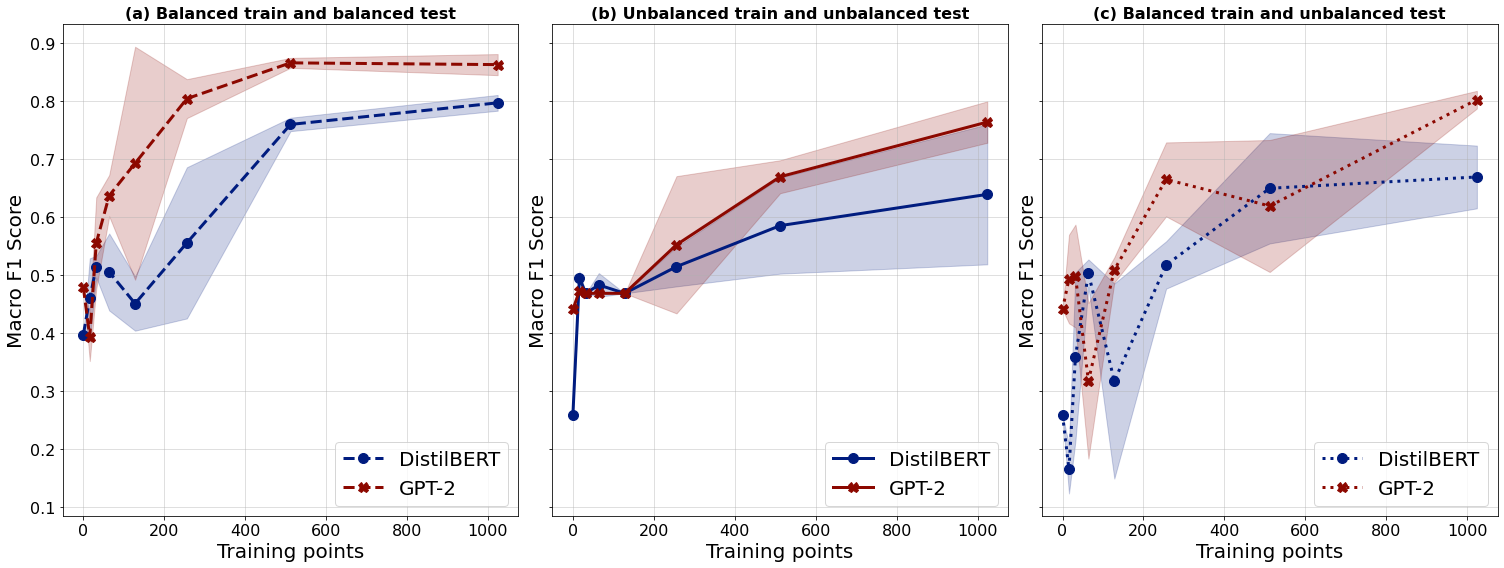}    \caption{\label{fig:binary_movie_sentiment_prompt_eng_f1}Binary movie sentiment classification results of prompt engineering varied by model.}
\end{figure}

Figure \ref{fig:binary_movie_sentiment_transfer_learning_f1} presents the Macro F1 scores of transfer learning for the binary movie sentiment classification task trained with DeBERTa-v3 and DistilBERT. Figure \ref{fig:binary_movie_sentiment_prompt_eng_f1} presents the Macro F1 scores of prompt engineering for the binary movie sentiment classification task trained with GPT-2 and DistilBERT, based on the prompt `Does this text contain negative sentiment?'. 
Similar to binary abuse classification task, transfer learning obtains the same learning pattern over labelling budgets, regardless of models. The performance of both models peaks at 128 training points and does not improve with more data. 

\begin{figure}[ht!]
    \centering
    \includegraphics[width=\textwidth]{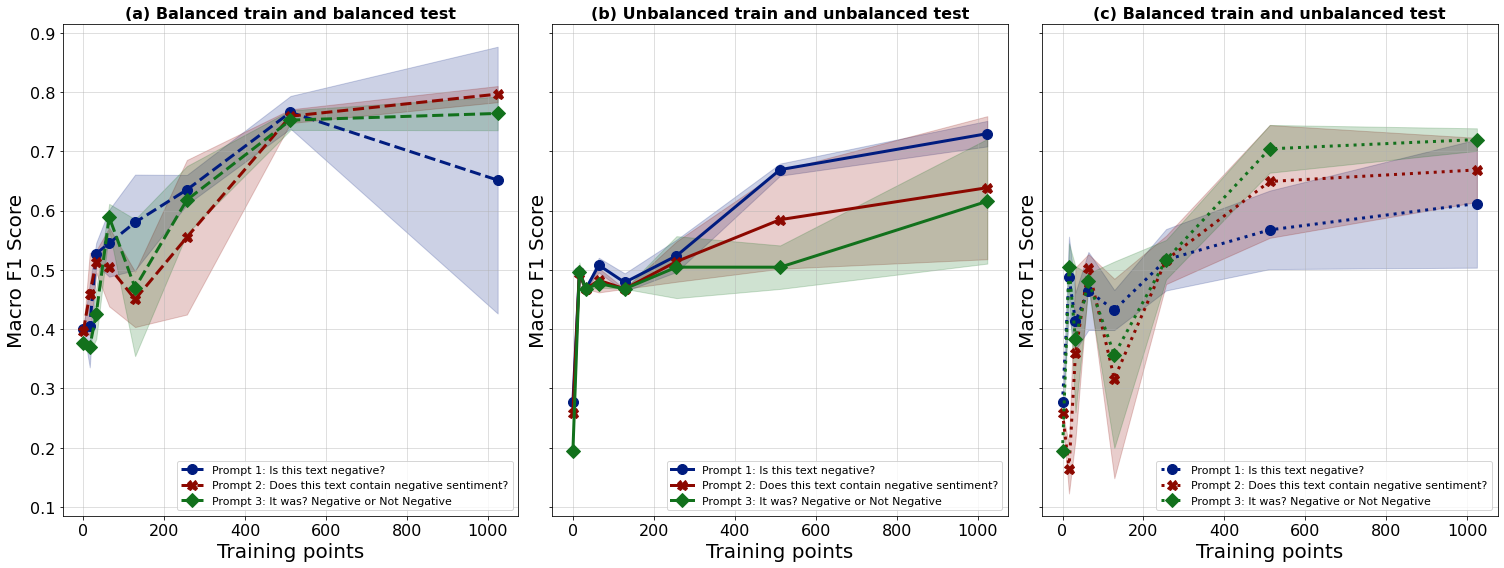}    \caption{\label{fig:binary_movie_sentiment_prompt_comparison_f1}Binary movie sentiment classification results of prompt engineering varied by template.}
\end{figure}

In Figure \ref{fig:binary_movie_sentiment_prompt_comparison_f1} we plot the Macro F1 scores for prompt engineering experiments with respect to the three different prompts used. Noticeably, high variance is seen in prompts at different training points. The prompt `Does this text contain negative sentiment?' shows high variance at small training, while `Is this text negative?' at large training. This confirms that the performance of prompt engineering is dependent on prompt selection.

In Table \ref{tab:results_tmdb} we present the results obtained by testing GPT-3, 3.5, and 4 in our TMDB dataset.

\begin{table}[htp!]
\begin{tabularx}{\textwidth}{llXl}
\toprule
Model & Number of points & Prompt & Macro F1 score \\
\toprule
GPT-3 & 855 & Using one word, does the movie review contain negative sentiment, Yes or No? & 0.67 \\
\addlinespace[5pt]
GPT-3 & 855 & Using one word, classify the sentiment of the movie review using 'Positive' or 'Negative'. & 0.85 \\
\addlinespace[5pt]
GPT-3 & 854  & You are a researcher who needs to classify movie reviews as containing negative sentiment or not containing negative sentiment. Using one word, does the movie review contain negative sentiment, Yes or No? & 0.11 \\
\addlinespace[5pt]
GPT-3.5 & 852 & Using one word, does the movie review contain negative sentiment, Yes or No? & 0.92 \\
\addlinespace[5pt]
GPT-3.5 & 843 & Using one word, classify the sentiment of the movie review using 'Positive' or 'Negative'. & 0.90    \\
\addlinespace[5pt]
GPT-3.5 & 851 & You are a researcher who needs to classify movie reviews as containing negative sentiment or not containing negative sentiment. Using one word, does the movie review contain negative sentiment, Yes or No? & 0.90 \\
\addlinespace[5pt]
GPT-4  & 855 & Using one word, does the movie review contain negative sentiment, Yes or No? & 0.93 \\
\addlinespace[5pt]
GPT-4 & 855 & Using one word, classify the sentiment of the movie review using 'Positive' or 'Negative'. & 0.94 \\
\addlinespace[5pt]
GPT-4 & 855 & You are a researcher who needs to classify movie reviews as containing negative sentiment or not containing negative sentiment. Using one word, does the movie review contain negative sentiment, Yes or No? & 0.88 \\
\bottomrule
\end{tabularx}
\caption{\label{tab:results_tmdb}Results of zero-shot testing the GPT-3, 3.5 and 4 models on the collected TMDB dataset to label positive and negative movie reviews.}
\end{table}

\newpage
\subsection{Analysis of the two tasks using Na\"ive Bayes classifier with different adjustments}
\label{ssec:naive_bayes}

The Na\"ive Bayes classifier \citep{10.1007/978-3-540-30549-1_43} is a probabilistic algorithm widely used for tasks like text classification due to its simplicity and computational efficiency. It operates on the assumption that features (e.g., words in a document) are conditionally independent given the class label, allowing for straightforward calculation of probabilities. However, the independence assumption is often violated in real-world data, particularly in text classification, where features tend to be highly correlated. This can result in biased probability estimates and reduced classification performance.

\cite{rennie_tackling_nodate} proposed a solution to address these limitations by introducing TF-IDF weighting and normalization into the Na\"ive Bayes framework. Their approach adjusts feature importance by emphasizing terms that are frequent in a document but not common across all classes, reducing the classifier's sensitivity to common words that contribute little to class distinction. Furthermore, they normalized feature vectors to mitigate the influence of document length, ensuring fair comparisons across texts of varying sizes.

\cite{fatourechi2008comparison} presented another refinement by incorporating logarithmic term frequency weighting. This adjustment reduces the disproportionate impact of extremely high term frequencies, which can dominate classification decisions despite potentially being less informative. By smoothing these term frequencies, their method makes the classifier more robust to outliers and improves its ability to generalize across different datasets.

\begin{figure}[ht]
    \centering
    \includegraphics[width=\textwidth]{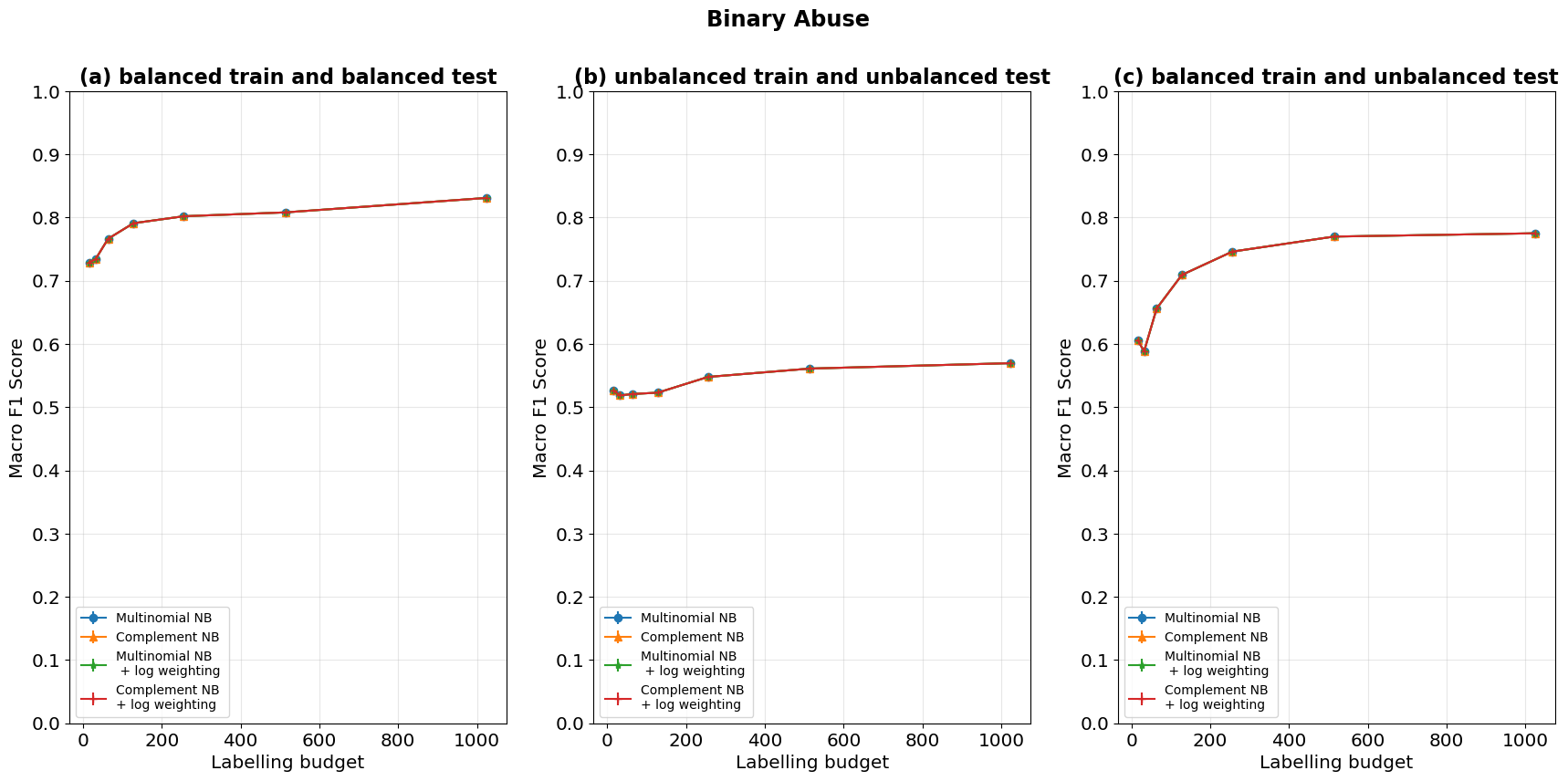}
    \caption{F1 macro scores when performing the Binary Abuse task with the four different Na\"ive Bayes.}
    \label{fig:NB_binary_abuse}
\end{figure}

\begin{figure}[ht]
    \centering
    \includegraphics[width=\textwidth]{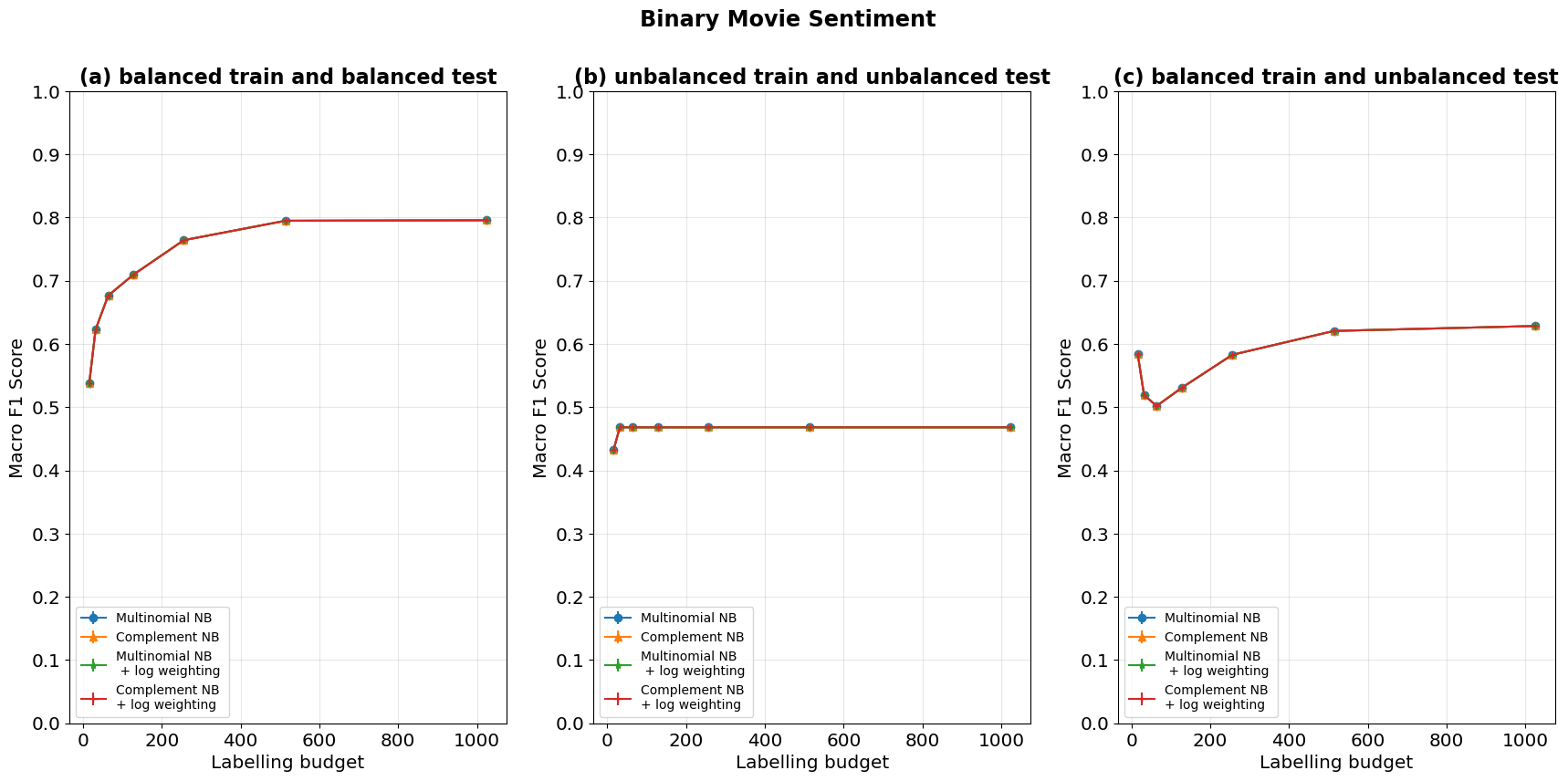}
    \caption{F1 macro scores when performing the Binary Movie Sentiment task with the four different Na\"ive Bayes.}
    \label{fig:NB_movie_sentiment}
\end{figure}

In this subsection we compare the Na\"ive Bayes classifiers with the two different fixes described above. We also combine these two into a single classifier, thus having a comprehensive exploration of this method. We end up with four different Na\"ive Bayes classifiers: the one used in the main body of the manuscript \cite{10.1007/978-3-540-30549-1_43} (called Multinomial NB here, which is the same as in the Python package \texttt{sklearn}), the one proposed by \cite{rennie_tackling_nodate} (called Complement NB), and each one of them with a log scaling as proposed by \cite{fatourechi2008comparison}. In Figures \ref{fig:NB_binary_abuse} and \ref{fig:NB_movie_sentiment} we see the F1-macro results for each one of the tasks.

\begin{figure}[ht]
    \centering
    \begin{subfigure}{0.45\textwidth}
        \includegraphics[width=\textwidth]{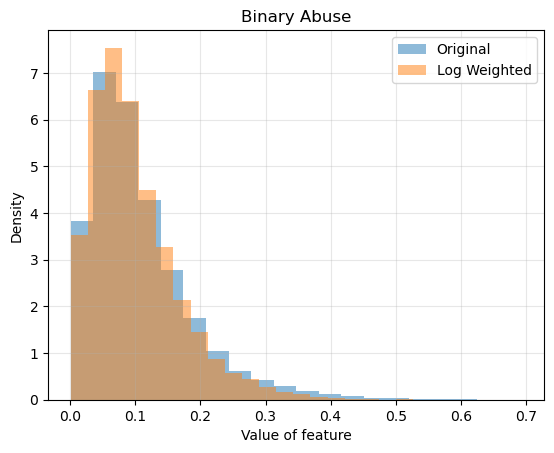}
        \caption{}
    \end{subfigure}
    \begin{subfigure}{0.45\textwidth}
        \includegraphics[width=\textwidth]{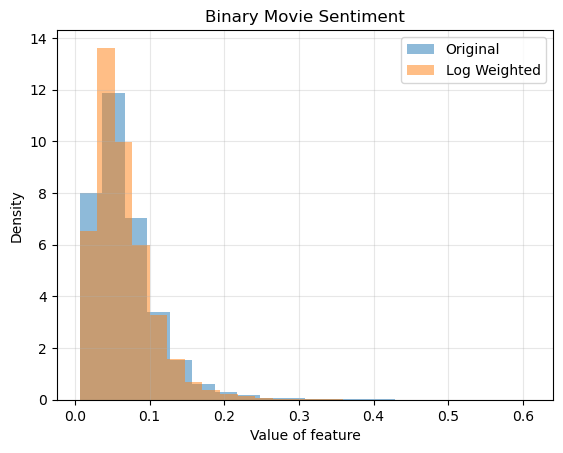}
        \caption{}
    \end{subfigure}
    \caption{Distribution of features of the vectorized training sets for each task, before and after applying a log scaling.}
    \label{fig:NB_dist_features}
\end{figure}

There is no distinction between the four different classifiers as can be seen in Figures \ref{fig:NB_binary_abuse} and \ref{fig:NB_movie_sentiment}. To understand this, we do two extra analyses. First, in order to understand why the log scaling proposed by \cite{fatourechi2008comparison} does not seem to have any effect, we perform a z-test between the two feature's distributions of the vectorized training sets before and after the log scaling. The histogram of the distribution for both tasks can be seen in Figure \ref{fig:NB_dist_features}. The z-scores are 0.625 and 0.421 for the binary abuse and binary movie sentiment tasks respectively, showing that the log scaling does not any statistically significant change. Thus, it is expected that we would not observe any difference between the Na\"ive Bayes classifiers with and without log scaling.

\begin{figure}[ht]
    \centering
        \includegraphics[width=\textwidth]{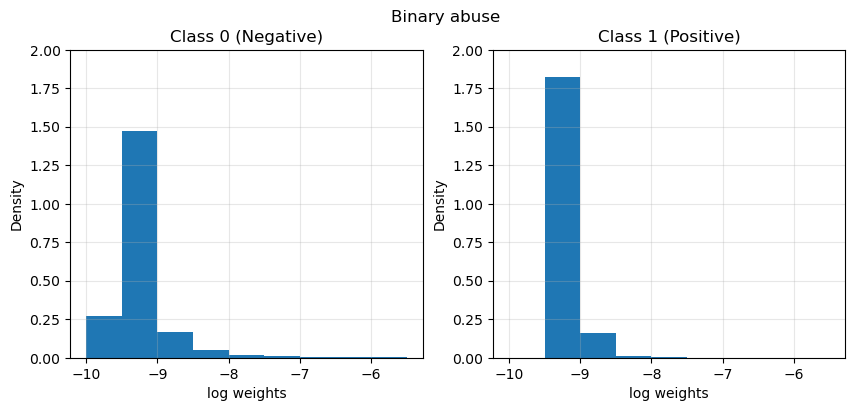}
        \caption{\label{fig:conditional_probabilities_binary_abuse} distribution of the weights for each feature of the trained multinomial Na\"ive Bayes classifier trained for the Binary Abuse task.}
    
\end{figure}

\begin{figure}[ht]
    \centering
        \includegraphics[width=\textwidth]{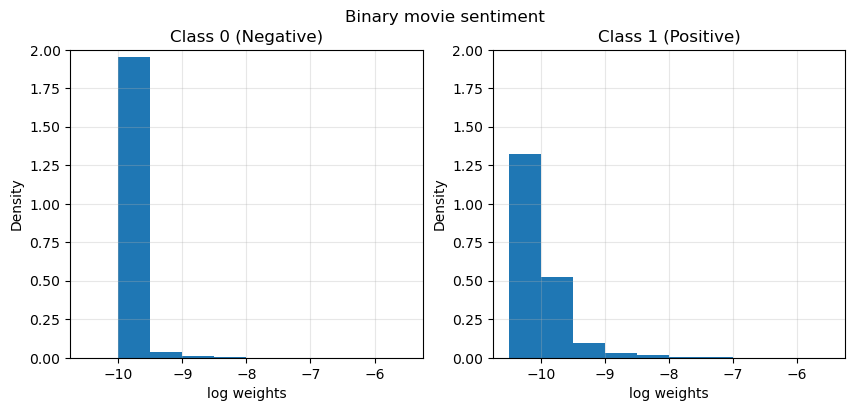}
        \caption{\label{fig:conditional_probabilities_binary_movie_sentiment} distribution of the weights for each feature of the trained multinomial Na\"ive Bayes classifier trained for the Binary Movie Sentiment task.}
    
\end{figure}

To understand why there is not any difference between the Multinomial NB classifier \citep{10.1007/978-3-540-30549-1_43} and the complement NB classifier \citep{rennie_tackling_nodate}, we look at the distribution of weights of the trained Multinomial NB classifier for each label. These can be seen in Figure \ref{fig:conditional_probabilities_binary_abuse}  for the Binary Abuse task and in Figure \ref{fig:conditional_probabilities_binary_movie_sentiment}  for the Binary Movie Sentiment task. The four distributions do not show any kind of imbalance, thus making the adjustment of the Complement NB, implemented after \cite{rennie_tackling_nodate}, without effect.

\clearpage

\subsection{Labelling Functions}
\label{ssec:labelling_functions}
In Table \ref{tab:labelling_functions} we present a list of all the Labelling Functions used for both classification tasks. 

    \begin{xltabular}{\textwidth}{XllXll}
        \toprule
        Task & Type & Label & Explanation & Coverage & Overalap\\
        \midrule
         Binary abuse & Keywords & ABUSE & If any of the words considered abusive from a given list of keywords is found in the text, label the text as abusive. & 0.36 & 0.36 \\ 
         Binary abuse & Heuristic & ABUSE & If the length of the text is above 9500 characters (limit of 10 000 characters), label the text as abusive. & 0.00 & 0.00 \\
         Binary abuse & NLP & ABUSE & If the polarity of the text is below -0.25, label the text as abusive. & 0.07 & 0.07  \\
         Binary abuse & NLP & NOT\_ABUSE & If the polarity of the text is equal or more than 0.1, label the text as non abusive. & 0.41 & 0.41  \\
         Binary abuse & Annotators & Both & Take the annotators' data and use it as labelling functions: given an annotator $i$ labelling a set of texts, the annotator-function labels the previously annotated texts and abstains with all non-annotated ones. & 0.02 & 0.02  \\
         Binary movie sentiment & Keywords & Positive & If any of the words considered related to positive sentiments is found from a given list of keywords, label the text as positive. & 0.90 & 0.90  \\
         Binary movie sentiment & Keywords & Negative & If any of the words considered related to negative sentiments is found from a given list of keywords, label the text as negative. & 0.62 & 0.62 \\
         Binary movie sentiment & Regex & Positive & If any expression related to the acting, the direction, the edition, the sound or in general to the film or the movie contains a positive keyword, then label the text as positive. & 0.25 & 0.25 \\
         Binary movie sentiment & Regex & Negative & If any expression related to the acting, the direction, the edition, the sound or in general to the film or the movie contains a negative keyword, then label the text as negative. & 0.03 & 0.03 \\
         Binary movie sentiment & NLP & Negative & If the polarity of the text is below -0.05, label it as negative. & 0.03 & 0.03 \\
         Binary movie sentiment & NLP & Negative & If the polarity of the text is above 0, label it as positive. & 0.95 & 0.95 \\
         Binary movie sentiment & NLP & Positive & If the subjectivity of the text is below 0.3, label it as positive. & 0.01 & 0.01 \\
         \bottomrule

    \caption{A description of the labelling functions used for both classification tasks.}
    \label{tab:labelling_functions}
        \end{xltabular}

\subsection{Zero-shot Prompt Engineering}
\label{ssec:zs_pe}
In Table \ref{tab:zs_pe} we present the results of the GPT-3.5-turbo for each of the prompts used. Some examples of non-response generated by GPT-3.5-turbo can be seen in \ref{tab:zs_pe2}.

    \begin{xltabular}{\textwidth}{lXlllX}
        \toprule
        Task & Prompt & GPT-3 & GPT-3.5 & GPT-4 & N Non-response \\
        \midrule
         Binary abuse & Using one word, does the internet comment contain toxic language, Yes or No? & 0.721 & 0.733 & 0.781 & 21\\ 
         Binary abuse & Using one word, is this internet comment using toxic language, Yes or No? & 0.667 & 0.684 & 0.797 & 13 \\
         Binary abuse & You are a researcher who needs to classify comments on the internet as containing abusive language or not containing abusive language. Using one word, does the internet comment contain abusive language, Yes or No? & 0.713 & 0.802 & 0.813 & 19 \\
         Binary movie sentiment & Using one word, does the movie review contain negative sentiment, Yes or No? & 0.952 & 0.923 & 0.961 & 2 \\
         Binary movie sentiment & Using one word, classify the sentiment of the movie review using 'Positive' or 'Negative'. & 0.941 & 0.945 & 0.964 & 8 \\
         Binary movie sentiment & You are a researcher who needs to classify movie reviews as containing negative sentiment or not containing negative sentiment. Using one word, does the movie review contain negative sentiment, Yes or No? & 0.876 & 0.918 & 0.939 & 7\\
         \bottomrule
             \caption{Results of the zero-shot prompts across models and prompts in terms of macro F1 score.}
    \label{tab:zs_pe}
    \end{xltabular}

\begin{table}[ht]
    \centering
    \begin{tabularx}{\textwidth}{lX}
        \toprule
        Task & Non-response \\
        \midrule
         Binary abuse & Cannot determine if the internet comment contains abusive language or not based on the given text \\ 
         Binary abuse & I cannot determine if this internet comment contains abusive language or not as it is not a complete sentence  \\
         Binary abuse & The internet comment does not contain a clear indication of abusive language  \\
         Binary abuse & The given internet comment does not contain any language  \\
         Binary abuse & I cannot determine if the internet comment contains abusive language or not as the comment is incomplete and ends abruptly  \\
         Binary movie sentiment & Neutral \\
         Binary movie sentiment & Mixed \\
         \bottomrule
    \end{tabularx}
    \caption{Non-responses generated by GPT-3.5-turbo}
    \label{tab:zs_pe2}
\end{table}

\end{document}